\documentclass[10pt,letterpaper]{article}

\usepackage[margin=1in]{geometry}
\usepackage[T1]{fontenc}
\usepackage{lmodern}
\usepackage{amsmath,amssymb,amsthm,mathtools}
\usepackage{booktabs,graphicx,microtype,placeins,tabularx,longtable,array}
\usepackage[authoryear,round]{natbib}
\setcitestyle{authoryear,round,citesep={;},aysep={,},yysep={;}}
\usepackage[hidelinks]{hyperref}
\hypersetup{
  pdftitle={Human Evaluation with Automatic Scores: Separating Sampling and Estimation Gains},
  pdfauthor={Yang Liu, Ran Zou, Zhengtao Gui, Chao Xue}
}

\newcommand{\Var}{\operatorname{Var}}
\newcommand{\Cov}{\operatorname{Cov}}

\makeatletter
\renewcommand{\maketitle}{%
  \begingroup
  \centering
  {\LARGE\bfseries \@title\par}
  \vspace{1.1em}
  {\normalsize \@author\par}
  \endgroup
  \vspace{0.8em}
}
\makeatother

\title{Human Evaluation with Automatic Scores:\\Separating Sampling and Estimation Gains}
\author{%
  Yang Liu\textsuperscript{1,\textdagger,*}\quad
  Ran Zou\textsuperscript{2,\textdagger}\quad
  Zhengtao Gui\textsuperscript{2}\quad
  Chao Xue\textsuperscript{3}\\[7pt]
  {\small
    \textsuperscript{1}Department of Statistics and Operations Research,\\
    University of North Carolina at Chapel Hill, USA\\[2pt]
    \textsuperscript{2}Department of Statistics, University of California, Irvine, USA\\[2pt]
    \textsuperscript{3}School of Computer Science and Engineering, University of New South Wales, Australia
  }\\[6pt]
  {\footnotesize
    \textsuperscript{\textdagger}Equal contribution.\qquad
    \textsuperscript{*}Corresponding author:
    \href{mailto:yangliu1@unc.edu}{\texttt{yangliu1@unc.edu}}
  }%
}
\date{}
\begin{document}
\raggedbottom
\maketitle
\begin{abstract}
Automatic judges offer inexpensive information about model outputs, but
an evaluation team must decide how to turn that information into fewer
human annotations. Scores can guide sampling, correct an estimate, or
support adaptive acquisition, and these uses may capture overlapping
gains. We compare their incremental value on fixed evaluation pools from
three preference datasets, and 19 attributed-question-answering systems. Matched comparisons use the same scores,
account for pilot labels and budget rules, and separate acquisition
from estimation. Score stratification is a strong reference: after
ten-bin stratification, the best fixed local linear correction offers
only 1.00--5.27\% further MSE reduction in the primary configurations.
Unequal allocation offers larger gains in some pools, but can increase
error when confident scores conceal variable human outcomes. Comparison on AttributedQA, shrinking learned quotas toward score-only quotas lowers pooled MSE for 17 of 19 systems relative
to the score-only procedure, but ordinary pooled 95\% intervals cover
only 88.09--93.54\%. The same shrinkage improves on the score-only reference in only one of eight preference configurations. Ranking
experiments further show that lower individual MSE
need not improve system selection. These findings support a simple
starting procedure, proportional sampling within score strata followed
by a stratified human mean, and identify what additional evidence is
needed to justify correction or adaptation for the intended evaluation
output.
\end{abstract}

\section{Introduction}
\label{sec:revision_intro}

An evaluation team can often score every model response automatically,
while the cost of human annotation limits how many responses people can
assess \citep{zheng2023judging,chaganty2018price}. \footnote{We used large language model tools to assist with manuscript editing, and code development. The authors are responsible for the context.} The challenge is to use
those scores to choose responses for annotation and estimate system
performance from the acquired labels. We study these choices on a fixed
set of evaluation questions and model responses. We assess each procedure
by how accurately it recovers the system scores and rankings that full
human annotation of that set would produce.

Existing methods use scores at both stages of this process.
Prediction-assisted evaluation combines automatic scores with human
measurements of their errors \citep{chaganty2018price}.
Prediction-powered inference (PPI) \citep{angelopoulos2023ppi} develops
this correction into a framework for point estimation and statistical
inference. Learning the prediction weight, as in
PPI++ \citep{angelopoulos2023ppipp}, adapts the correction to the
information the predictions provide. With few labels, however, fitting
that weight can introduce substantial variability, motivating
regularized regression \citep{eyre2025regression}. Post-hoc calibration
also allows the relationship between scores and outcomes to be
nonlinear \citep{vanderlaan2026calibeating}.
Score-based stratified estimators have also been studied in Bayesian
PPI, including on AttributedQA~\citep{hofer2024bayesian}.
Scores can instead, or additionally, guide acquisition.
StratPPI \citep{fisch2024stratppi} studies stratification and allocation
alongside prediction correction. Active statistical inference
\citep{zrnic2024active} directs annotation toward items where predictions
are uncertain, and robust sampling addresses inaccuracies in the
uncertainty estimates used for acquisition \citep{li2025robust}.
These developments make both the estimator and the human sample part
of the design choice.

For a team choosing a procedure, the practical question is whether
additional fitting or adaptation is worthwhile when a simpler design
already uses the same scores. One can divide responses into score
strata, allocate labels in proportion to stratum size, and combine the
sampled human averages using population weights
\citep{cochran1977sampling,sarndal1992model}. This stratified human mean
requires neither a fitted correction nor a pilot. Prior model-evaluation work on computer vision has separated stratification, sampling, and estimation and shown how stratification can reduce
the additional gain from prediction correction
\citep{fogliato2024framework}.
We ask what further gains correction, unequal allocation, and pilot
adaptation provide beyond this reference at a given human-label budget in LLM evaluations.
We also ask which component produces an improvement and whether
combining components preserves the benefits they provide separately.

A paid pilot makes the cost and attribution problems concrete. It
commits part of the label budget before the final allocation is chosen
and requires another annotation round. Its outcomes can affect both
which labels are collected next and how the pilot labels are weighted
in the final estimate. This dependence can introduce bias when all
acquired labels are pooled, an established issue in adaptive stratified
sampling \citep{salehi2010efficient}. A favorable result for the complete
procedure therefore does not by itself identify the benefit of learning
quotas. We separate these effects by holding the estimator fixed when
changing quotas and holding the acquired labels fixed when changing
estimators. Every pilot label counts toward the total budget.
For model-assisted estimators, design-based analyses characterize how
within-stratum residual variation determines efficient allocation
\citep{mozer2026stratified}. In practice, a paid pilot must estimate
the variation relevant to the chosen estimator from limited labels.
Small pilots can therefore produce unstable allocations
\citep{cai2024smallpilots}, so we test whether regularizing the learned
quotas improves their additional value.

We apply these comparisons to released human judgments from
AlpacaFarm \citep{dubois2023alpacafarm},
Chatbot Arena \citep{chiang2024arena}, and
HelpSteer3 \citep{wang2025helpsteer3}, together with 19
AttributedQA (AQA) systems \citep{bohnet2022attributed}.
Each replay reveals only the human outcomes requested by a procedure.
The complete outcomes define the evaluation target and support exact
risk calculations for fixed-coefficient estimators under fixed sampling
designs, as well as retrospective oracle diagnostics. This distinguishes
the improvement available in principle from what a procedure recovers
with limited labels. The comparisons retain each dataset's annotation
unit, distinguish exact, capped, and expected label expenditure, and
report score production separately from human-label costs.

The results explain why these additional steps have different value
across evaluation pools. With ten score bins per natural group, the best
fixed local linear correction offers only 1.00--5.27\% further
mean-squared error (MSE) reduction in the primary configurations.
Coarser partitions leave more room for correction, showing that
stratification and correction can use overlapping information.
Unequal allocation can produce larger gains because it addresses a
further question: where do human outcomes vary? It can also increase
error when confident predictions direct too few labels to strata with
substantial human variation. On AQA, regularized pilot allocation provides a positive case for
adaptive point estimation, but its benefit depends on how pilot
labels enter the estimate. The importance of the reference
is also visible in the active-method comparison: on AQA, our finite-pool
adaptation of RobustASI \citep{li2025robust} improves on its same-pilot
uniform control for 18 of 19 systems, but on the proportional
stratified mean for four.

These MSE comparisons are only part of the evaluation. A leaderboard
depends on errors in differences between systems, including the
covariance induced by how their samples are selected. A confidence
interval depends on the estimate's bias and the reliability of its
estimated standard error. Our ranking and coverage experiments show
that lower individual MSE can coexist with worse best-system selection
or insufficient coverage. Annotation savings must therefore be assessed
against the output that the evaluation is intended to preserve as the
budget falls.

Our contributions are:
\begin{itemize}
    \item \textbf{Controlled comparisons of the additional value of
    scores.} We compare correction, allocation, and pilot adaptation
    against a proportional score-stratified human mean. The comparisons
    vary acquisition and estimation separately, charge pilot labels to
    the budget, and account for differences in expenditure rules.

    \item \textbf{Mechanistic explanations of gains and failures.}
    Oracle calculations and controlled comparisons connect performance
    to remaining within-stratum signal, coefficient-fitting noise,
    inaccurate allocation proxies, and pilot-label reuse. Regularized
    alternatives test whether the failure of one learned quota rule
    persists after its allocations are moderated.

    \item \textbf{Evaluation of the outputs that labels are purchased
    to support.} Budget curves, joint ranking experiments, and coverage
    diagnostics test whether mean-estimation gains translate into
    better system selection and uncertainty reporting. These results
    inform deployment choices and provide a reproducible protocol for
    assessing proposed improvements.
\end{itemize}

\section{The Evaluation Task and the Comparisons}\label{sec:design}
\subsection{The evaluation problem}

We evaluate how different uses of automatic scores improve human evaluation estimates under a limited annotation budget.
Our aim is to separate gains from choosing which items to label
from gains from combining the acquired labels with scores.
We measure estimation accuracy through mean-squared error, then ask
whether improvements also support reliable system rankings and
confidence intervals.

\paragraph{Evaluation tasks and data.}
We use datasets with released human annotations, which allow us to
compare each estimate with the result obtained from the complete
evaluation pool.
Table~\ref{tab:pools} summarizes the primary datasets.
AlpacaFarm~\citep{dubois2023alpacafarm}, Chatbot Arena~\citep{chiang2024arena}, and HelpSteer3~\citep{wang2025helpsteer3} support preference evaluation:
the target is the average human preference for a designated response
across the evaluated pairs.
Arena also supports a contrast between two opponent-group means.
AttributedQA (AQA)~\citep{bohnet2022attributed} supports attribution evaluation: the target is
the fraction of a system's answers supported by their cited evidence.
Its 19 systems answer the same 1,000 questions, allowing us to evaluate
both individual system estimates and the rankings they produce.

\begin{table}[htbp]
\centering\small
\caption{Primary evaluation pools. Preference scores average both response
orders. One acquired label reveals one released outcome, which can
aggregate several human judgments. Appendix~\ref{app:protocol} specifies
annotation units; Appendix~\ref{app:cost} reports scoring resources and
cost definitions.}
\label{tab:pools}
\begin{tabular}{lrlp{4.8cm}}
\toprule
Pool & Items & Human target & Automatic score \\
\midrule
AlpacaFarm & 648 & Preference & Mistral and Qwen judges \\
Chatbot Arena & 2,870 & Preference; contrast
  & Same judges; contrast uses 1,247 items \\
HelpSteer3 & 2,017 & Preference
  & Same judges; four task groups \\
AttributedQA & 1,000/system & Attribution; ranking
  & Released AutoAIS; 19 systems \\
\bottomrule
\end{tabular}
\end{table}

\paragraph{Automatic scores.}
Each pool pairs human outcomes with scores available for every item.
The preference datasets use Mistral and Qwen as judges
\citep{psr8_mistral_small32,qwen2026model},
and AQA uses released AutoAIS scores~\citep{bohnet2022attributed}.
These sources provide the auxiliary information whose use we compare.

For each preference pair, the judge assigns probabilities to three
outcomes: the first response is better, the responses are tied, or
the second response is better.
For Mistral, we normalize the model's next-token probabilities over
the three judgment labels.
For Qwen, we prompt the model to report the three probabilities
directly as numerical values.
Either output is converted into a score for the designated response:
its win probability plus half the tie probability.
LLM judgments can depend on which response is presented first~\citep{zheng2023judging}. To reduce this sensitivity, we score each pair in both orders, express both scores relative to the same designated response, and average them.
The resulting scores are shared by all methods within a matched comparison.

\paragraph{Controlled label acquisition.}
In each replay, a procedure receives the automatic scores and
permitted grouping information for the entire evaluation pool.
Human annotations are hidden and revealed only for requested items.
Every acquired label, including those used in a pilot, counts
toward the annotation budget.
We compare the resulting estimates with reference quantities
computed from the complete human annotations, repeating acquisition
to measure sampling variability.
Individual experiments specify their budget rules; scoring costs
are reported separately in Appendix~\ref{app:cost}.

\paragraph{Separating the sources of improvement.}
The experiments proceed from using existing scores to acquiring
additional information, and finally to evaluating the decisions
supported by the estimates.

Section~\ref{sec:scores} examines what the available scores can achieve
before any pilot is collected.
We first compare human-mean estimates with and without score
stratification, which groups items with similar automatic scores.
Within a fixed partition, we apply alternative estimators to the
same acquired samples to isolate the additional value of score-based
correction.
We then keep the partition and human-mean estimator fixed while
changing the label quotas across strata.
This separates the effect of allocating labels from the effect of
correcting the estimate.

Section~\ref{sec:adaptivity} asks whether observing pilot labels helps
choose better allocations.
We compare fixed and pilot-learned allocation rules using shared
pilot samples and the same total label budget.
For each allocation, alternative estimators use the same acquired
labels, allowing us to separate the value of learning quotas from
the consequences of reusing pilot observations.
We also compare ASI~\citep{zrnic2024active} and RobustASI~\citep{li2025robust} with uniform-acquisition controls
within their workflows and with a score-stratified human mean.
The former comparison measures the benefit of selective acquisition;
the latter evaluates the complete procedure against a simpler use
of the same scores, with budget differences accounted for explicitly.

Section~\ref{sec:decisions} asks whether lower mean-squared error
translates into better evaluation decisions.
On AQA, we compare ranking performance across label budgets and
examine how sharing questions between systems affects the comparison.
For confidence intervals, we distinguish errors caused by biased
point estimates from those caused by unstable variance estimates.
These experiments test whether a procedure's estimation gains carry
through to the ranking or uncertainty statement that an evaluator
intends to report.

\subsection{From sampled human labels to evaluation estimates}

\paragraph{The population and the target.}
Let $U=\{1,\ldots,N\}$ denote the complete set of evaluation items
for one target. Each item has a human outcome $y_i$ and an automatic
score $s_i$.
For the preference datasets, $y_i$ records human preference for the
designated response. For AQA, $U$ contains one system's answers and
$y_i$ records whether an answer is supported by its evidence.
In each case, the target is the full-pool human average,
\[
    \theta=\frac{1}{N}\sum_{i\in U}y_i.
\]

The AQA ranking experiment in Section~\ref{sec:decisions} compares
these means across systems. The Arena contrast, defined below,
compares two group means.

\paragraph{Estimating the target from a stratified sample.}
We begin with the static sampling designs used in
Section~\ref{sec:scores}.
If the dataset has natural groups, such as task categories or
opponent families, let $G_g$ denote group $g$ and
$N_g=|G_g|$ its size.
Without such groups, we treat the complete pool $U$ as one group.
Within each group, we partition items by automatic score so that
items with similar scores belong to the same stratum.
We index the resulting strata across all groups as
$U_1,\ldots,U_H$, with $N_h=|U_h|$, and write
$\mathcal H_g$ for the indices of strata within group $g$.

Let $L$ be the total human-label budget.
The quota $n_h$ specifies how many items receive human labels
in stratum $h$, with $\sum_{h=1}^{H}n_h=L$.
These quotas are fixed before observing human outcomes.
Within each stratum, we draw a simple random sample
$S_h\subseteq U_h$ of size $n_h$, without replacement and
independently across strata.
The sampled human average is
$\overline y_{S_h}=n_h^{-1}\sum_{i\in S_h}y_i$.

To estimate the full-pool human mean, we combine these sampled
averages in two steps.
Within each natural group, we weight each stratum by its share
of that group's population.
We then weight each group estimate by the group's share of
the complete pool:
\begin{equation}
\begin{aligned}
    \widehat\theta_g
    &=\sum_{h\in\mathcal H_g}
      \frac{N_h}{N_g}\overline y_{S_h},\\
    \widehat\theta_{\rm mean}
    &=\sum_g\frac{N_g}{N}\widehat\theta_g
      =\sum_{h=1}^{H}a_h\overline y_{S_h},
      \qquad a_h=\frac{N_h}{N}.
\end{aligned}
\label{eq:main_mean}
\end{equation}
We call this the \emph{stratified human mean}, following the classical stratified expansion estimator
\citep{cochran1977sampling,sarndal1992model}.
The two levels of weighting reduce to a single stratum weight
because $(N_g/N)(N_h/N_g)=N_h/N$.
We use the compact expression with $a_h$ throughout the following
comparisons.

Each sampled average $\overline y_{S_h}$ is unbiased for the
full-stratum human average $\overline y_{U_h}$.
Since the target satisfies
$\theta=\sum_h a_h\overline y_{U_h}$,
Equation~\eqref{eq:main_mean} is unbiased for the full-pool mean
under this fixed sampling design.
The weights reflect population sizes, even when groups or strata
receive disproportionate shares of the label budget.
Automatic scores therefore contribute through the sampling design,
while the averages used to estimate human quality contain only
the acquired human outcomes.

The Arena contrast uses the same weighted construction to estimate
$\theta=\overline y_{G_+}-\overline y_{G_-}$, where $G_+$ and $G_-$
are the two opponent groups being compared. For example, let $G_+$ contain battles against Claude-family models and $G_-$ contain battles against GPT-3.5-family models.
In both groups, $y_i$ records human preference for the GPT-4-family response, with a win, tie, and loss scored as $1$, $1/2$, and $0$.
The contrast then measures how much higher or lower GPT-4's average human preference score is against Claude than against GPT-3.5 on the observed battle pools.
For strata inside $G_+$, set $a_h=N_h/|G_+|$;
for strata inside $G_-$, set $a_h=-N_h/|G_-|$.
With these signed weights, the compact expression
$\sum_h a_h\overline y_{S_h}$ estimates the difference
between the two group means.
Appendix~\ref{app:estimators} specifies the contrast.

\paragraph{Using scores to correct the estimate.}
Stratification uses scores to organize the sample.
Prediction-assisted estimation also uses their numerical values
after the labels have been acquired.
Let $\overline s_{U_h}$ be the average score over the complete
stratum and $\overline s_{S_h}$ the average over its sampled items.
The former is known because every item has an automatic score.
A score-corrected estimator is
\begin{equation}
    \widehat\theta_b
    =\sum_{h=1}^{H}a_h
    \left\{
        \overline y_{S_h}
        +b_h\left(\overline s_{U_h}-\overline s_{S_h}\right)
    \right\}.
    \label{eq:fixed}
\end{equation}
The correction accounts for differences between the sampled items
and the full stratum that are visible in their scores.
For example, if the sampled items have unusually high scores,
a positive coefficient $b_h$ adjusts their human average downward.

Setting $b_h=0$ recovers the stratified human mean.
Setting $b_h=1$ gives fixed prediction-error correction:
the full-stratum score average is corrected by the average
human-minus-score residual among sampled items.
Estimating $b_h$ from acquired labels allows the strength of the
correction to adapt to the observed relationship between scores
and human outcomes.
These choices are finite-pool versions of regression and
prediction-assisted estimators
\citep{sarndal1992model,angelopoulos2023ppi,angelopoulos2023ppipp}.
Under the fixed sampling design above, coefficients chosen
independently of the acquired outcomes give unbiased estimates.
Fitting coefficients on the same labels can affect both bias
and variance.

These estimators provide the two controlled comparisons in
Section~\ref{sec:scores}.
Applying Equations~\ref{eq:main_mean} and~\ref{eq:fixed} to the
same samples isolates the contribution of prediction correction.
Keeping Equation~\eqref{eq:main_mean} fixed while changing the quotas
isolates the contribution of allocation.
When quotas depend on pilot outcomes, label reuse becomes an
additional design choice.
Section~\ref{sec:adaptivity} therefore defines the corresponding
two-stage and active estimators separately.
For its item-level formulas, we use the equivalent target weights
$\omega_i=Na_h/N_h$ for $i\in U_h$; these equal one for a pool mean.

\paragraph{Measuring improvement against a common reference.}
For any estimator $\widehat\theta$, its mean-squared error is
\[
    R
    =\mathbb{E}\!\left[(\widehat\theta-\theta)^2\right]
    =\operatorname{Var}(\widehat\theta)
     +\left(\mathbb{E}[\widehat\theta]-\theta\right)^2.
\]
The expectation covers repeated label acquisition and any fitting
based on those labels. The complete human outcomes and automatic
scores remain fixed.

Our default reference uses score stratification, proportional
label allocation, and the stratified human mean in
Equation~\eqref{eq:main_mean}.
For a pool mean, proportional allocation assigns labels in
proportion to stratum sizes, so $n_h\approx LN_h/N$.
For the signed contrast, allocation follows $|a_h|$.
Integer rounding, minimum counts, and capacity constraints are
specified in Appendix~\ref{app:allocation}.
Writing $\widehat\theta_{\rm mean}^{\rm prop}$ for the estimate
under this reference design, we define
\[
    R_0
    =\mathbb{E}\!\left[
        \left(\widehat\theta_{\rm mean}^{\rm prop}-\theta\right)^2
      \right].
\]

We report $R/R_0$, so a ratio of $0.9$ means a 10\% reduction in
MSE relative to this reference.
The original-group mean, which uses the existing task or opponent
groups without further score stratification, is included as a
separate comparator.
This makes the benefit of introducing score strata visible
alongside the additional benefits of correction or unequal quotas.
Comparisons using other references identify them explicitly.
Section~\ref{sec:decisions} then evaluates ranking errors and
interval coverage directly, since these outcomes depend on more
than the MSE of an individual estimate.

\section{What Can the Available Scores Already Do?}\label{sec:scores}
Automatic scores can improve evaluation through both sampling and
estimation. We separate these contributions in two steps.
First, we examine the benefit of score stratification and ask whether
prediction correction further improves estimates on the same samples.
We then keep the stratified human-mean estimator fixed and change
the allocation of labels across strata.
The primary comparisons use $L=\lfloor N/4\rfloor$ labels and ten
score bins within each natural group.
Coarser partitions and smaller budgets test how these choices affect
the conclusions.

\subsection{Correction after stratification: how much information remains?}
\label{sec:correct}

A score can help organize sampling even when its numerical value
is an imperfect prediction of the human outcome.
If scores separate items with different human means, sampling within
score strata can reduce the variability caused by an uneven sample.
The stratified human mean in Equation~\eqref{eq:main_mean} uses this
information through a classical sampling design
\citep{sarndal1992model}.
We use it as the reference and apply alternative estimators to the
same acquired samples.
Holding the partition, quotas, and samples fixed isolates what
prediction correction adds after stratification.

\paragraph{Correction rules and why we compare them.}
In our setting, PPI corrects an automatic prediction average using the prediction
errors observed on labeled items~\citep{angelopoulos2023ppi}.
PPI++ additionally learns how strongly to weight the predictions
\citep{angelopoulos2023ppipp}, while StratPPI allows that weight
to vary across strata~\citep{fisch2024stratppi}.
These approaches motivate our comparison between fixed correction,
$b_h=1$ in Equation~\eqref{eq:fixed}, and fitted correction.
We evaluate the local fit with one coefficient per score stratum, and
also include a group-shared control with one coefficient across
the score strata within each natural group.
The latter tests whether estimating fewer coefficients improves
performance when individual strata contain few labels.

Let $\widehat C_{ys,h}$ be the sample covariance of human outcomes
and scores in stratum $h$, and let $S^2_{s,h}$ be the known score
variance over the complete stratum $h$.
For our finite-pool comparisons, the local fit uses the full-stratum score variance, matching the local point estimator implemented in
GLIDE~\citep{martinon2026glide}. We also construct a group-shared coefficient using the same variance weights:
\begin{equation}
 \widehat b_h=\frac{\widehat C_{ys,h}}{S^2_{s,h}},
 \qquad
 \widehat b_g^{\rm shared}
 =\frac{\sum_{h\subset \mathcal H_g}c_h\widehat C_{ys,h}}
        {\sum_{h\subset \mathcal H_g}c_hS^2_{s,h}}
\label{eq:coefficients}
\end{equation}
where $c_h=a_h^2\frac{1-n_h/N_h}{n_h}$. Here, $\mathcal H_g$ is the set of stratum indices within
natural group $g$, as defined above.
The weights $c_h$ account for each stratum's contribution to
estimation variance, including its sampling fraction.
A zero denominator sets the corresponding coefficient to zero.

Fitting a correction coefficient can itself introduce substantial
variability when labels are scarce~\citep{eyre2025regression}.
We therefore compare the unrestricted fits with two forms of
regularization.
Clipping restricts fitted coefficients to $[0,1]$, a restriction
also considered in PPI++~\citep{angelopoulos2023ppipp};
we use it here as a sensitivity check on extreme fitted values.
Ridge-PPI instead penalizes large regression coefficients,
following \citet{eyre2025regression}.
The group-shared adaptation uses sample score variances and
a positive penalty:
\begin{equation}
 \widehat b_g^{\rm ridge}(\alpha)
 =\frac{\sum_{h\subset \mathcal H_g}c_h\widehat C_{ys,h}}
        {\sum_{h\subset \mathcal H_g}c_h(\widehat S^2_{s,h}+\alpha)}.
 \label{eq:ridge_main}
\end{equation}
The penalty is selected by leave-one-out prediction error using only
acquired labels. Coefficient sharing, clipping, and ridge regularization
test whether stabilizing the fitted coefficients improves estimation.
The main table reports group-shared linear and Ridge fits.
Table~\ref{tab:coefficient_controls} in the appendix compares local
and group-shared coefficients, with and without clipping, on the
same sampled labels. The implementation and tuning details are
given in Appendix~\ref{app:r38_correction}.

\paragraph{What could correction achieve with known coefficients?}
The coefficient-sharing, clipping, and Ridge comparisons above
test whether stabilizing coefficient estimates improves correction.
We next examine how much improvement is available when the
coefficients are known. To do so, we compute the best fixed
coefficients from the complete evaluation pool, keeping the
partition and quotas unchanged. This oracle measures the remaining
scope for linear correction after stratification and helps explain
why even a stable fit may offer little additional benefit.

For fixed coefficients, define $e_i=y_i-b_hs_i$ for $i\in U_h$.
The corrected estimator is unbiased under the stratified sampling
design, and its exact MSE is
\begin{equation}
 R_b=\sum_h c_h S^2_{e,h},
 \label{eq:fixedrisk}
\end{equation}
where $S^2_{e,h}$ is the full-stratum residual variance.
The factor $1-n_h/N_h$ in $c_h$ accounts for sampling without
replacement. Setting every coefficient to zero recovers the stratified
human mean, whose risk under these same quotas is
$R_{\rm mean}=\sum_h c_hS^2_{y,h}$.
Here, $S^2_{y,h}$ is the variance of the human outcomes
over the complete stratum $U_h$, with divisor $N_h-1$.
Under the proportional allocation used for this correction
comparison, $R_{\rm mean}=R_0$.

Let $C_{ys,h}$ denote the covariance between human outcomes and
scores in the complete stratum. Minimizing the fixed-coefficient
risk separately in each stratum gives
\begin{equation}
 b_h^\star=\frac{C_{ys,h}}{S^2_{s,h}},
 \qquad
 R_{\rm mean}-R_{b^\star}
 =\sum_{h:S^2_{s,h}>0}
   c_h\frac{C_{ys,h}^2}{S^2_{s,h}}.
 \label{eq:r38-oracle-local}
\end{equation}
We set $b_h^\star=0$ when the stratum's score variance is zero.
The coefficients $b^\star$ define the oracle estimator
$\widehat\theta_{b^\star}$, and $R_{b^\star}$ denotes its risk.
Because zero correction is among the available choices,
$R_{b^\star}\leq R_{\rm mean}$.

The difference $R_{\rm mean}-R_{b^\star}$ therefore measures the
MSE reduction available to fixed linear correction after
stratification. A small difference means that even supplying
the best coefficients leaves little room for improvement.
This calculation uses complete human outcomes retrospectively
and measures association within strata. A strong association
across the full pool can coexist with little remaining linear
association within its score strata.

\begin{table}[p]
\centering\small
\setlength{\tabcolsep}{2.7pt}
\renewcommand{\arraystretch}{1.06}
\caption{\textbf{Separating gains from stratification, correction, and allocation.}
Primary-budget MSE divided by the matching ten-bin proportional-mean risk
(reference $1$; lower is better). 
\emph{Mean} sets $b_h=0$: scores guide sampling, while estimation uses
weighted human averages without correction. 
Parentheses give one MCSE from 10,000 paired draws; means and oracles
are exact. Among ten-bin non-oracle columns, \textbf{bold} marks displayed
ratios $\leq0.95$ or $\geq1.05$, highlighting improvement and deterioration.
\textit{Italics} identify the full-label oracle.
M/Q denote Mistral/Qwen; AQA uses AutoAIS.}
\label{tab:correction_r38}
\begin{tabular}{llrrrrrr}
\toprule
& & \shortstack{No score\\bins}
& \multicolumn{3}{c}{Ten bins: proportional allocation}
& \multicolumn{2}{c}{Ten bins: working allocation} \\
\cmidrule(lr){4-6}\cmidrule(lr){7-8}
Target / system & Score & \shortstack{Original\\mean}
& Linear & Ridge & \textit{Oracle} & \shortstack{Mean\\($b=0$)} & Ridge \\
\midrule
\multicolumn{8}{l}{\textbf{A. Preference targets}} \\
Alpaca & M & 1.225 & 1.009\,{\scriptsize(0.0014)} & 1.005\,{\scriptsize(0.0010)} & \textit{0.987} & 0.996 & 1.000\,{\scriptsize(0.0011)} \\
Alpaca & Q & 1.404 & 1.005\,{\scriptsize(0.0022)} & 1.002\,{\scriptsize(0.0014)} & \textit{0.965} & 1.028 & 1.032\,{\scriptsize(0.0012)} \\
\addlinespace[2pt]
Arena mean & M & 1.072 & 1.008\,{\scriptsize(0.0018)} & 1.003\,{\scriptsize(0.0014)} & \textit{0.990} & 0.998 & 1.000\,{\scriptsize(0.0013)} \\
Arena mean & Q & 1.163 & 1.002\,{\scriptsize(0.0016)} & 1.001\,{\scriptsize(0.0012)} & \textit{0.986} & 1.039 & 1.040\,{\scriptsize(0.0011)} \\
Arena contrast & M & 1.071 & 1.008\,{\scriptsize(0.0022)} & 1.004\,{\scriptsize(0.0018)} & \textit{0.984} & 1.003 & 1.006\,{\scriptsize(0.0018)} \\
Arena contrast & Q & 1.159 & 1.010\,{\scriptsize(0.0017)} & 1.005\,{\scriptsize(0.0013)} & \textit{0.978} & 1.027 & 1.031\,{\scriptsize(0.0010)} \\
\addlinespace[2pt]
HelpSteer3 & M & 1.424 & 1.002\,{\scriptsize(0.0025)} & 1.002\,{\scriptsize(0.0019)} & \textit{0.967} & 0.970 & 0.970\,{\scriptsize(0.0022)} \\
HelpSteer3 & Q & 1.622 & 1.006\,{\scriptsize(0.0025)} & 1.004\,{\scriptsize(0.0017)} & \textit{0.985} & \textbf{0.929} & \textbf{0.929}\,{\scriptsize(0.0014)} \\
\midrule
\multicolumn{8}{l}{\textbf{B. AttributedQA: all 19 complete systems}} \\
LLM-as-retriever & AutoAIS & 2.412 & 0.990\,{\scriptsize(0.0031)} & 0.995\,{\scriptsize(0.0025)} & \textit{0.970} & 0.986 & 0.986\,{\scriptsize(0.0015)} \\
Post + rerank & AutoAIS & 1.671 & 1.003\,{\scriptsize(0.0022)} & 1.004\,{\scriptsize(0.0016)} & \textit{0.966} & \textbf{1.717} & \textbf{1.717}\,{\scriptsize(0.0008)} \\
Post-1 & AutoAIS & 2.256 & 1.001\,{\scriptsize(0.0026)} & 1.000\,{\scriptsize(0.0019)} & \textit{0.985} & \textbf{0.929} & \textbf{0.929}\,{\scriptsize(0.0012)} \\
Post-2 & AutoAIS & 2.077 & 1.002\,{\scriptsize(0.0028)} & 1.004\,{\scriptsize(0.0022)} & \textit{0.979} & 1.038 & 1.038\,{\scriptsize(0.0015)} \\
Post-3 & AutoAIS & 2.264 & 1.000\,{\scriptsize(0.0028)} & 1.002\,{\scriptsize(0.0022)} & \textit{0.965} & 0.992 & 0.991\,{\scriptsize(0.0009)} \\
Post-4 & AutoAIS & 2.188 & 1.006\,{\scriptsize(0.0021)} & 1.002\,{\scriptsize(0.0015)} & \textit{0.985} & \textbf{1.092} & \textbf{1.094}\,{\scriptsize(0.0006)} \\
Post-5 & AutoAIS & 2.812 & 0.978\,{\scriptsize(0.0034)} & 0.985\,{\scriptsize(0.0029)} & \textit{0.958} & \textbf{0.690} & \textbf{0.679}\,{\scriptsize(0.0017)} \\
Post-6 & AutoAIS & 2.069 & 1.009\,{\scriptsize(0.0017)} & 1.004\,{\scriptsize(0.0012)} & \textit{0.983} & \textbf{1.061} & \textbf{1.062}\,{\scriptsize(0.0010)} \\
Post-7 & AutoAIS & 2.963 & 1.010\,{\scriptsize(0.0019)} & 1.003\,{\scriptsize(0.0011)} & \textit{0.978} & \textbf{0.674} & \textbf{0.674}\,{\scriptsize(0.0006)} \\
Post-8 & AutoAIS & 2.195 & 0.969\,{\scriptsize(0.0039)} & 0.975\,{\scriptsize(0.0034)} & \textit{0.947} & \textbf{0.950} & \textbf{0.938}\,{\scriptsize(0.0021)} \\
\addlinespace[2pt]
RTR + rerank & AutoAIS & 1.479 & 1.003\,{\scriptsize(0.0020)} & 1.002\,{\scriptsize(0.0014)} & \textit{0.957} & \textbf{1.513} & \textbf{1.509}\,{\scriptsize(0.0016)} \\
RTR-10 & AutoAIS & 1.742 & 1.005\,{\scriptsize(0.0019)} & 1.003\,{\scriptsize(0.0015)} & \textit{0.978} & 0.954 & 0.954\,{\scriptsize(0.0013)} \\
RTR-11 & AutoAIS & 2.574 & 0.991\,{\scriptsize(0.0027)} & 0.995\,{\scriptsize(0.0024)} & \textit{0.966} & \textbf{0.902} & \textbf{0.894}\,{\scriptsize(0.0018)} \\
RTR-12 & AutoAIS & 1.831 & 0.988\,{\scriptsize(0.0029)} & 0.993\,{\scriptsize(0.0024)} & \textit{0.963} & 0.982 & 0.974\,{\scriptsize(0.0022)} \\
RTR-2 & AutoAIS & 2.743 & 0.988\,{\scriptsize(0.0036)} & 0.996\,{\scriptsize(0.0032)} & \textit{0.953} & \textbf{0.505} & \textbf{0.505}\,{\scriptsize(0.0005)} \\
RTR-4 & AutoAIS & 2.202 & 0.977\,{\scriptsize(0.0035)} & 0.983\,{\scriptsize(0.0030)} & \textit{0.958} & \textbf{0.913} & \textbf{0.911}\,{\scriptsize(0.0016)} \\
RTR-6 & AutoAIS & 2.248 & 0.980\,{\scriptsize(0.0034)} & 0.986\,{\scriptsize(0.0029)} & \textit{0.957} & \textbf{0.933} & \textbf{0.929}\,{\scriptsize(0.0017)} \\
RTR-8 & AutoAIS & 1.903 & 0.996\,{\scriptsize(0.0030)} & 0.997\,{\scriptsize(0.0024)} & \textit{0.963} & \textbf{1.061} & \textbf{1.054}\,{\scriptsize(0.0019)} \\
RTR-9 & AutoAIS & 2.103 & 1.002\,{\scriptsize(0.0022)} & 1.003\,{\scriptsize(0.0015)} & \textit{0.968} & \textbf{0.781} & \textbf{0.779}\,{\scriptsize(0.0012)} \\
\bottomrule
\end{tabular}
\end{table}

\paragraph{What remains after score stratification?}
Table~\ref{tab:correction_r38} includes all primary preference
configurations and all 19 AQA systems.
Its proportional-allocation columns use the matching stratified
human mean as the reference value of 1.
The ``Original mean'' column uses the same total budget $L$ but
samples within the natural groups without dividing them into
score bins. Let $S_g^{\rm orig}$ be a simple random sample of
$n_g$ items drawn without replacement from group $G_g$,
independently across groups. With target weight $w_g$ for group
$g$, the estimator and allocation are
\[
\begin{aligned}
\widehat\theta_{\rm orig}
&=\sum_g w_g\left(
    \frac{1}{n_g}\sum_{i\in S_g^{\rm orig}}y_i
  \right),\\
n_g
&\approx L\frac{|w_g|}{\sum_{g'}|w_{g'}|},
\qquad \sum_g n_g=L.
\end{aligned}
\]
For a pool mean, $w_g=N_g/N$, so $n_g\approx LN_g/N$.
For the Arena contrast, $(w_+,w_-)=(1,-1)$, giving the difference
between two group sample means and approximately half the labels
in each group. Integer rounding, minimum counts, and capacity
constraints follow Appendix~\ref{app:allocation}.

For each AQA system there is one group, so this reduces to
\[
\widehat\theta_{\rm orig}
=\frac{1}{L}\sum_{i\in S^{\rm orig}}y_i,
\]
where $S^{\rm orig}$ is an SRS of $L$ answers from that system's
pool. The table reports
$\mathbb{E}[(\widehat\theta_{\rm orig}-\theta)^2]/R_0$.
Thus, a value above 1 means that introducing score strata reduces
MSE at the same total label budget.

To check whether this gain comes from informative scores, we
repeat the stratification comparison after removing their
relationship to human outcomes. In one control, scores are
randomly reassigned among items within each natural group,
preserving the group's score distribution. In the other, all
scores are set to a constant and random tie breaking determines
the bins. We rebuild ten bins per group for each repetition
and retain the human outcomes, total label budget, and
proportional allocation rule.

Across 100 repetitions per configuration, both controls give
average stratified-mean MSE close to the Original mean's MSE
(Appendix~\ref{app:Information-free}). This supports attributing the observed sampling
gain to how the scores separate items with different human
outcomes.

Read the remaining columns against the ten-bin proportional mean, whose risk ratio is 1. The Linear, Ridge \citep{eyre2025regression}, and Oracle columns measure what correction adds after stratification. The working-allocation
columns instead change the quotas and are examined in the Section~\ref{sec:Allocation}.

After ten-bin stratification, the local oracle reduces MSE by
1.00--3.48\% on the preference configurations and 1.46--5.27\%
on AQA.
Thus, even with the best fixed coefficient supplied, the remaining
linear gains are small relative to the same-quota mean.
Shared Ridge stays close to that reference, with MSE ratios of
1.001--1.005 on preferences and 0.975--1.004 on AQA.
Some AQA systems benefit from correction, but the additional
improvements are smaller than those obtained by introducing
score strata.

\paragraph{Does nonlinear correction change this account?}
Limited opportunity for linear correction could still leave
a nonlinear relationship between scores and human outcomes.
We examine this possibility using the published Calibeating PPI
implementation~\citep{vanderlaan2026calibeating}.
LinearCal, MonoSpline, and IsoCal fit a calibration map
$\widehat f_g(s)$ within each natural group using acquired
labels and inverse-inclusion weights.
The corrected estimate combines the calibrated full-pool scores
with the observed human residuals:
\begin{equation}
 \widehat\theta_f
 =\sum_g\sum_{h\subset \mathcal H_g}a_h\left\{
 \frac{1}{N_h}\sum_{i\in U_h}\widehat f_g(s_i)
 +\frac{1}{n_h}\sum_{i\in S_h}
   [y_i-\widehat f_g(s_i)]\right\}.
 \label{eq:nonlinear_main}
\end{equation}
We compare these calibration rules on the same samples within
each partition and retain their upstream defaults
(Appendix~\ref{app:sensitivity}).
Varying the partition at the same total label budget then tests
whether the value of calibration depends on how much information
the sampling design already uses.

\begin{table}[t]
\centering
\small
\setlength{\tabcolsep}{7pt}
\renewcommand{\arraystretch}{1.08}
\caption{
\textbf{Calibration gains depend on how finely scores are stratified.}
Results for all eight preference configurations at the primary budget.
Each block uses its own matching proportional stratified mean as the
MSE reference $1$; lower is better.
Entries are ranges across configurations; MSE ranges are bold throughout.
MCSE gives the range of Monte Carlo standard errors for the MSE ratios.
}
\label{tab:nonlinear_r38}
\begin{tabular}{lrr}
\toprule
Calibration & \textbf{MSE ratio} & MCSE \\
\midrule
\multicolumn{3}{l}{
    \textbf{1 bin/group: no score stratification}
} \\
\addlinespace[2pt]
IsoCal
& \textbf{0.648--0.965}
& {\scriptsize 0.0045--0.0078} \\
LinearCal
& \textbf{0.616--0.945}
& {\scriptsize 0.0018--0.0080} \\
MonoSpline
& \textbf{0.622--0.944}
& {\scriptsize 0.0032--0.0082} \\
Local correction
& \textbf{0.618--0.946}
& {\scriptsize 0.0019--0.0080} \\
\midrule
\multicolumn{3}{l}{\textbf{5 bins/group}} \\
\addlinespace[2pt]
IsoCal
& \textbf{0.995--1.030}
& {\scriptsize 0.0032--0.0067} \\
LinearCal
& \textbf{0.979--1.000}
& {\scriptsize 0.0006--0.0025} \\
MonoSpline
& \textbf{0.983--1.005}
& {\scriptsize 0.0022--0.0034} \\
Local correction
& \textbf{1.000--1.022}
& {\scriptsize 0.0046--0.0067} \\
\midrule
\multicolumn{3}{l}{\textbf{10 bins/group}} \\
\addlinespace[2pt]
IsoCal
& \textbf{1.000--1.033}
& {\scriptsize 0.0027--0.0069} \\
LinearCal
& \textbf{0.992--1.005}
& {\scriptsize 0.0003--0.0014} \\
MonoSpline
& \textbf{0.994--1.003}
& {\scriptsize 0.0014--0.0022} \\
Local correction
& \textbf{1.027--1.086}
& {\scriptsize 0.0066--0.0089} \\
\bottomrule
\end{tabular}
\end{table}

Table~\ref{tab:nonlinear_r38} shows that dependence.
With one bin per natural group, all three calibration rules
improve MSE across the eight preference configurations.
With ten bins, their ratios lie close to one.
The overlap between stratification and correction therefore
extends to these nonlinear controls: a finer score partition
can leave less additional benefit for calibration.

\paragraph{Deployment implication.}
Evaluate correction against a human mean that already uses the
same scores for stratification.
Coarse strata can leave useful signal for a fitted adjustment,
while fine strata can leave little signal and few labels per fit.
The partition and correction should therefore be chosen together
at the intended annotation budget.
For example, at 100 labels, ten bins per group create 40 strata
in HelpSteer3/Qwen, 30 of which receive only two labels.
Five bins give lower human-mean MSE in this setting (Appendix~\ref{app:small_budget}).
This illustrates why increasing the number of strata and adding
a more flexible correction do not automatically improve evaluation.

\subsection{Allocation: which strata need more human labels?}
\label{sec:Allocation}
\label{sec:r45_allocation}

Section~\ref{sec:correct} asked what score-based correction adds after stratification.
We now ask whether scores can improve how the label budget is distributed
across those strata. We first keep the partition, budget, and stratified
human mean fixed, and change only the quotas. We then add correction to
examine whether the two uses of scores work well together.

\paragraph{What should guide the quotas?}
More labels are useful where human outcomes remain variable.
Let $S_{y,h}$ be the standard deviation of all human outcomes in stratum $h$.
For quotas $\mathbf n=(n_1,\ldots,n_H)$, the exact MSE of the stratified
human mean is
\[
R_{\rm mean}(\mathbf n)
=\sum_h a_h^2 S_{y,h}^2\left(\frac{1}{n_h}-\frac{1}{N_h}\right),
\qquad
S_{y,h}^2=\frac{1}{N_h-1}\sum_{i\in U_h}(y_i-\overline y_{U_h})^2.
\]
This expression accounts for both a stratum's target weight and its human
variation. With continuous quotas and inactive bounds, minimizing it gives
the classical Neyman allocation \citep{cochran1977sampling}:
\[
n_h^{\rm Neyman}
=L\frac{|a_h|S_{y,h}}{\sum_j|a_j|S_{y,j}}.
\]
Proportional allocation follows $n_h\propto |a_h|$ instead.
Because $S_{y,h}$ is unknown before acquiring labels, using this principle
requires a useful approximation from the available judge outputs.

\paragraph{A working variance matched to the annotation protocol.}
StratPPI constructs allocation weights from predicted outcome
distributions \citep{fisch2024stratppi}.
We adapt this construction to each dataset's annotation protocol.
For each item, the judge outputs specify a working distribution
for the numerical human outcome that will be recorded.
The existing score $s_i$ is the mean of this distribution.
We denote its variance by $u_i$, which describes the predicted
uncertainty in that item's human outcome.

The distinction matters because the same mean can imply different
uncertainty. A predicted tie with probability one and equally likely
win and loss outcomes both give $s_i=1/2$, but their variances are
$u_i=0$ and $u_i=1/4$, respectively.
To predict variation across human outcomes within a stratum, we
combine this within-item uncertainty with variation in the predicted
means across items:
\begin{equation}
v_h^{\rm work}
=\underbrace{\frac{1}{N_h}\sum_{i\in U_h}u_i}_{\text{uncertainty within items}}
 +\underbrace{\frac{1}{N_h}\sum_{i\in U_h}(s_i-\overline s_{U_h})^2}_{\text{variation across items}},
\qquad
n_h^{\rm work}\propto |a_h|\sqrt{v_h^{\rm work}}.
\label{eq:working}
\end{equation}
The two terms have different roles: uncertainty about an individual human
outcome and differences in predicted outcomes across items both create
variation within a stratum.

For binary AQA labels, $u_i=s_i(1-s_i)$.
For Arena and HelpSteer3, the judge's win, tie, and loss probabilities
$(p_{W,i},p_{T,i},p_{L,i})$ describe values $(1,1/2,0)$, giving
$s_i=p_{W,i}+p_{T,i}/2$ and $u_i=p_{W,i}+p_{T,i}/4-s_i^2$.
Alpaca instead averages $r_i$ binary human votes. Its working model uses
$u_i=s_i(1-s_i)/r_i$, assuming conditionally independent votes with mean
$s_i$. Appendix~\ref{app:r45_variance} gives the derivations, actual vote
counts, and sensitivity to dependence between votes. 

The working allocation uses no pilot labels. Both allocation rules satisfy
$2\leq n_h\leq N_h$ and spend exactly $L$ labels through the same bounded
rounding routine (Appendix~\ref{app:allocation}).
In Table~\ref{tab:correction_r38}, \emph{Mean} sets $b_h=0$ in
Equation~\eqref{eq:fixed}: it combines sampled human averages using their
target weights, with no score correction. Its comparison with the
proportional-mean reference $R_0$ therefore isolates allocation.

\emph{Ridge} uses the same working quotas and adds the regularized,
group-shared linear correction from Section~\ref{sec:correct} \citep{eyre2025regression}.
The acquired labels fit coefficients that adjust the final estimate;
the probabilities used for allocation remain unchanged. Mean and Ridge
thus distinguish allocation alone from allocation with fitted correction.
Table~\ref{r6:tab:extra_combinations} reports fixed $b_h=1$
correction under both quota rules and the unregularized
group-shared fit under working quotas.
The corresponding group-shared fit under proportional quotas
appears in Table~\ref{tab:correction_r38}.

\paragraph{Why the same allocation principle can help or harm.}
Working allocation lowers mean-estimator MSE for 13 of 19 AQA systems,
with effects ranging from a 49.5\% reduction to a 71.7\% increase.
Among the preference configurations, HelpSteer3 has MSE ratios of
$0.970$ for Mistral and $0.929$ for Qwen at the primary budget.
Alpaca's ratios are $0.996$ and $1.028$; Arena's range from $0.998$ to
$1.039$. Useful gains therefore depend on the evaluation pool and judge.

\begin{figure}[htbp]
\centering
\includegraphics[width=0.8\linewidth]{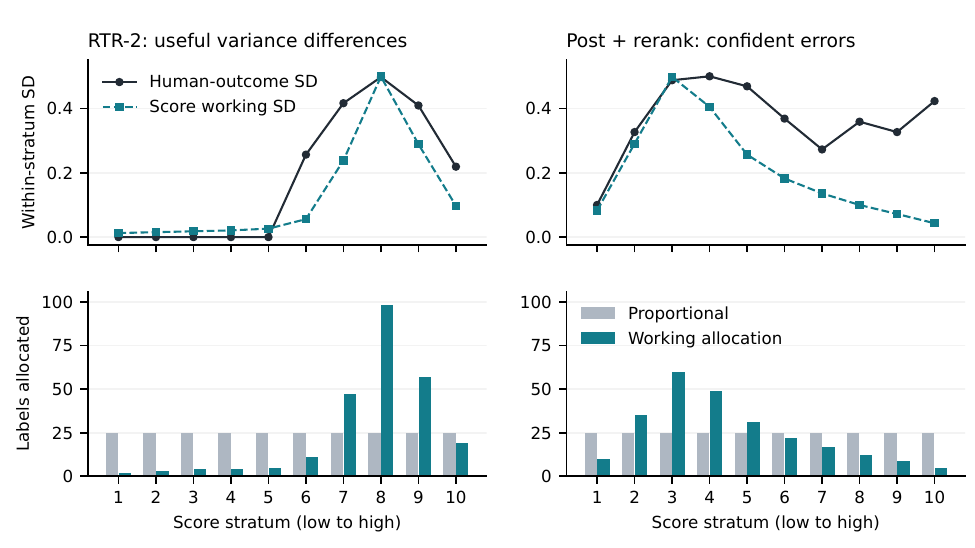}
\caption{\textbf{Score-based uncertainty helps when it tracks human variation.}
Two AQA systems at $L=250$ labels each. The upper panels compare
human-outcome SD $S_{y,h}$ with working SD $\sqrt{v_h^{\rm work}}$;
the lower panels show proportional and working quotas. Strata are ordered
by score. Complete human outcomes explain these results retrospectively.}
\label{fig:mechanism}
\end{figure}

Figure~\ref{fig:mechanism} explains the allocation mechanism through
two AQA systems. For RTR-2, the lowest five score strata have zero
human-outcome variance. Moving labels out of those strata makes more
labels available where outcomes differ. For Post + reranking, the
highest-score stratum has a human positive rate of $0.77$ and an SD of
$0.423$. Its almost-unit scores nevertheless imply little working
variance, reducing its quota from 25 labels to five. The rule assigns
too few labels to a stratum that still contributes substantial uncertainty.

\paragraph{How much improvement was available?}
A small gain could mean that proportional quotas were already close to
optimal, or that the working rule missed a useful opportunity.
Let $R_{\rm int}$ be the lowest mean-estimator MSE attainable by integer
quotas with the same partition, budget, and bounds. This diagnostic oracle
uses complete human outcomes. It optimizes quotas, whereas Section~\ref{sec:correct}'s
oracle optimized correction coefficients at fixed quotas.
Writing $R_{\rm work}=R_{\rm mean}(\mathbf n^{\rm work})$ and
$R_0=R_{\rm mean}(\mathbf n^{\rm prop})$, we obtain
\begin{equation}
\underbrace{\frac{R_0-R_{\rm work}}{R_0}}_{\text{realized MSE reduction}}
=\underbrace{\frac{R_0-R_{\rm int}}{R_0}}_{\text{available reduction}}
-\underbrace{\frac{R_{\rm work}-R_{\rm int}}{R_0}}_{\text{loss from chosen quotas}}.
\label{eq:opportunity_main}
\end{equation}
All three quantities use the same reference $R_0$.
For HelpSteer3/Qwen, optimal feasible quotas could reduce MSE by $11.09\%$.
The working quotas leave $3.96$ percentage points unused, delivering
the observed $7.13\%$ reduction. If the loss from the chosen quotas
exceeds the available reduction, allocation increases MSE.
Appendix~\ref{app:allocation} describes the exact integer optimization
\citep{wright2014allocation}.

\paragraph{What happens when judge confidence changes?}
StratPPI notes that miscalibrated confidence can make its allocation
heuristic overly aggressive \citep{fisch2024stratppi}.
We test this mechanism by applying a fixed temperature to the predicted
outcome probabilities \citep{guo2017calibration}:
\[
p_{ik,T}=\frac{p_{ik}^{1/T}}{\sum_j p_{ij}^{1/T}},
\qquad T\in\{0.5,1,2\}.
\]
Here $k$ indexes the possible outcomes: two for an AQA label or an
individual Alpaca vote, and three for a win, tie, or loss.
$T=0.5$ concentrates probability on the more likely outcomes, $T=1$
leaves the probabilities unchanged, and $T=2$ makes them more uniform.
For example, for a binary probability of $0.8$, the transformed values are
approximately $0.941$ and $0.667$ at $T=0.5$ and $T=2$.

Sharpening to $T=0.5$ increases MSE in 26 of 27 configurations.
Softening to $T=2$ reduces MSE in 20 and moves quotas closer to
proportional allocation in 25.
Quota distance is the fraction of label slots that must move between
strata to turn one allocation into the other. For example, changing
$(70,30)$ to $(50,50)$ moves 20 of 100 slots, giving distance $0.20$.
Appendix~\ref{app:r45_temperature} defines this metric and gives all
temperature results.

Better allocation need not imply better mean calibration. We measure
calibration by the size-weighted absolute gaps between stratum mean
scores and mean human outcomes. One preference configuration has lower
MSE at $T=2$ despite a larger calibration error. Softening can therefore
help by moderating quotas even when this measure of calibration worsens.

\paragraph{Deployment implication.}
Start with a variance model for the annotation actually being purchased,
then compare its quotas with proportional quotas using the same estimator
and budget. The proxy must locate strata with variable human outcomes;
overall judge accuracy does not answer that question. Independent
historical labels can help assess this property on similar evaluation
pools. Budget also matters: HelpSteer3/Qwen's allocation gain falls from
$7.13\%$ at the primary budget to $2.57\%$ at $L/N=5\%$ (Appendix~\ref{app:small_budget}).
The next section asks whether a paid pilot on the current pool supplies
enough additional information to improve this decision.

\section{When Is a Paid Pilot Worthwhile?}\label{sec:adaptivity}
Score-only allocation can miss the places where humans disagree with a
judge. A pilot offers a way to find those places before spending the rest
of the annotation budget. The practical question is whether the learned
allocation repays that investment. Answering it requires two comparisons:
how to use the labels already purchased, and whether their outcomes help
choose better final quotas. We first separate these decisions, then compare
the resulting designs with established active-inference procedures.

\subsection{Separate learning quotas from reusing pilot labels}
\label{sec:dif_est}

We keep the score strata and total label budget $L$ fixed.
A proportional pilot uses $P=\max\{\lceil L/5\rceil,2H\}$
labels, where $H$ is the number of strata.
In stratum $h$, let $A_h$ contain the $m_h$ pilot items,
sampled uniformly without replacement, with $\sum_hm_h=P$.
The final sample $T_h$ contains $q_h$ items drawn uniformly
without replacement from the remaining $M_h=N_h-m_h$ items.
Write $A=\bigcup_h A_h$ and $T=\bigcup_h T_h$ for the
complete pilot and final samples.
All reported designs use integer final counts satisfying
$2\le q_h\le M_h$ and $\sum_hq_h=L-P$.
The lower bound allows estimation of within-stratum variance
from the final sample.
Every pilot label counts toward the total budget.

Let $n_h^0$ denote the full-budget one-stage proportional
count in stratum $h$, with $\sum_h n_h^0=L$.
These counts are determined before observing human outcomes
from shares $|a_h|/\sum_j|a_j|$, using the one-stage
baseline's bounds and integer rounding.
They provide the reference allocation against which we
evaluate the two-stage procedures.

For a given quota rule, we compare three estimators on
the same acquired labels: Completion, Composite, and Pooling.
Writing $\bar y_{A,h}$ and $\bar y_{T,h}$ for the pilot
and final sample means, respectively, these estimators are
\begin{align}
 C_h
 &=
 \frac{m_h\bar y_{A,h}+M_h\bar y_{T,h}}{N_h},
 &
 \widehat\theta_C
 &=
 \sum_h a_h C_h,
 \nonumber\\
 \widehat\theta_B
 &=
 \sum_h a_h
 \left\{
 \lambda_h\bar y_{A,h}+(1-\lambda_h)C_h
 \right\},
 &
 \lambda_h
 &=
 \frac{m_h(N_h-n_h^0)}{n_h^0M_h},
 \nonumber\\
 \widehat\theta_P
 &=
 \sum_h a_h
 \frac{m_h\bar y_{A,h}+q_h\bar y_{T,h}}{m_h+q_h}.
 \label{eq:pilot_estimators_main}
\end{align}
where $\widehat\theta_C$, $\widehat\theta_B$, and
$\widehat\theta_P$ denote Completion, Composite, and
Pooling, respectively.

\paragraph{Completion: an unbiased reference under adaptive quotas.}
Completion applies the sampling-expansion principle to the
part of each stratum outside the pilot
\citep{sarndal1992model}.
It includes the observed pilot total directly and uses
$M_h\bar y_{T,h}$ to estimate the remaining total.
Conditional on the pilot, the final sample is uniform within
that remainder.
Completion therefore remains conditionally unbiased even
when pilot outcomes determine the final quotas.
It lets us evaluate allocation without introducing bias
through label reuse.

\paragraph{Composite: recover the baseline with an unbiased combination.}
The pilot mean $\bar y_{A,h}$ also estimates the full
stratum mean.
Motivated by fixed-weight combinations in two-wave estimation
\citep{yamin2026pilot}, Composite combines it
with $C_h$ using weights determined before observing pilot
outcomes.
We choose $\lambda_h$ in
Equation~\eqref{eq:pilot_estimators_main} so that, when
$q_h=n_h^0-m_h$, the combination equals the mean of all
$n_h^0$ acquired labels in that stratum.
Keeping these weights fixed under adaptive quotas preserves
unbiasedness over repeated pilot and final sampling.
This estimator tests whether a loss under Completion can
be reduced through a different unbiased use of the same labels.

\paragraph{Pooling: average all acquired labels within each stratum.}
Pooling gives every acquired label the same weight within
its stratum.
It is unbiased when both pilot and final counts are fixed
before observing outcomes.
Under adaptation, however, pilot outcomes can affect $q_h$
and hence their own weight in the final average.
This dependence can introduce bias, a recognized issue in
adaptive stratified sampling \citep{salehi2010efficient}.
We assess Pooling through both MSE and bias, and examine
its interval coverage separately.

We begin with the matched proportional control,
$q_h=n_h^0-m_h$.
As intended, Composite and Pooling recover the one-stage
stratified mean exactly.
Completion has about 19\% higher MSE at the primary budget
(Appendix~\ref{r6:app:pilot_risk}).
This comparison attributes the loss to the estimator's
weighting: the same two-stage samples recover the baseline
when combined differently.

We compare three final-quota rules under the same
pilot and total budget.
The proportional rule retains the matched control above.
The score-only rule applies the working allocation from
Section~\ref{sec:scores} to the remaining budget $L-P$.
Using the working variance $v_h^{\rm work}$ from
Equation~\eqref{eq:working}, its unconstrained target counts are
\[
    \widetilde q_h^s
    =
    (L-P)
    \frac{M_h|\omega_h|\sqrt{v_h^{\rm work}}}
    {\sum_j M_j|\omega_j|\sqrt{v_j^{\rm work}}},
\]
where $\omega_h=Na_h/N_h$ is the per-item target weight.
We write $q_h^s$ for the resulting integer final-stage
counts, which satisfy $\sum_hq_h^s=L-P$ and
$2\le q_h^s\le M_h$.
These counts depend only on the available judge outputs
and predetermined stratum and pilot counts, so they can
be computed before collecting any human labels.
The learned rule estimates the within-stratum variability of the pilot
residuals $y_i-s_i$. Its unconstrained target counts satisfy
\[
    \widetilde q_h \propto
    M_h|\omega_h|\bigl(\widehat S_{y-s,h}\bigr)^{1/2},
\]
where $\widehat S_{y-s,h}$ is the sample standard deviation of those
residuals. Taking its square root moderates allocation differences
driven by a small pilot. If any estimated residual standard deviation
is zero, the rule uses weights $M_h|\omega_h|$ for all strata.
We use this as a heuristic pilot-based allocation baseline, motivated
by residual-based allocation
\citep{fisch2024stratppi,mozer2026stratified} and small-pilot
regularization \citep{cai2024smallpilots}.
The target counts are converted into integer quotas satisfying the
bounds and total budget above. Appendix~\ref{app:r38_pilot}
gives the implementation details.

Applying all three estimators under each quota rule separates
the two decisions.
Comparisons within a quota rule measure the effect of
combining labels differently.
Comparisons across quota rules with the estimator held fixed
measure the effect of allocating the final labels differently.

\begin{figure}[htbp]
\centering
\includegraphics[width=0.75\linewidth]{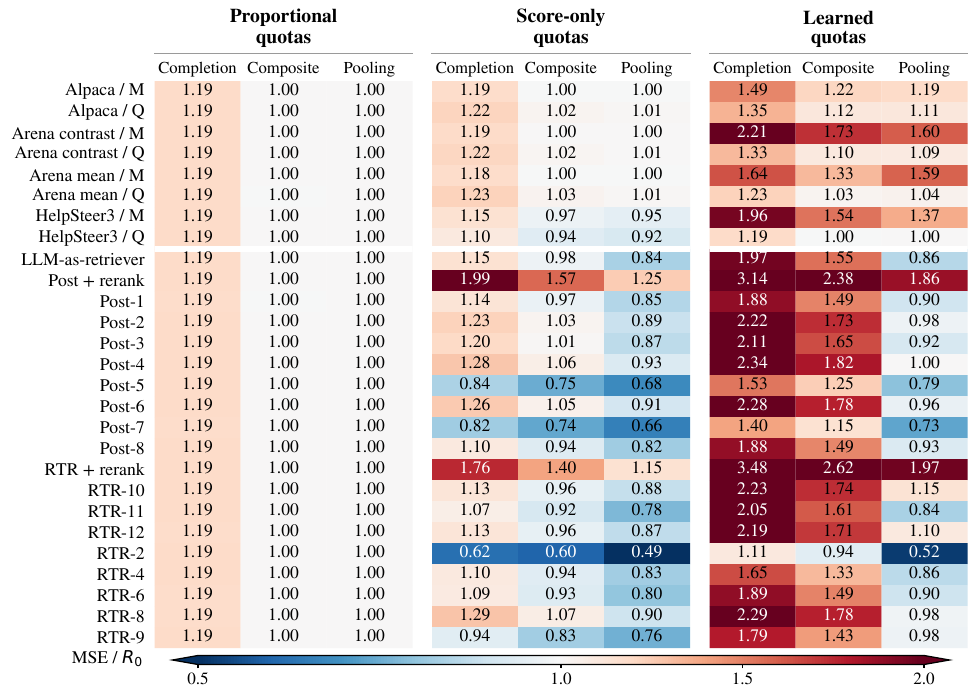}
\caption{
\textbf{Learning quotas and reusing labels are separate decisions.}
MSE relative to the one-stage proportional mean, using the
same pilot samples and total budgets within each configuration.
M and Q denote Mistral and Qwen.
Blue indicates lower MSE.
Within each group of three columns, the acquired labels
are identical and only the estimator changes.
}
\label{fig:pilot_matrix}
\end{figure}

Figure~\ref{fig:pilot_matrix} first compares estimators
under the same learned quotas.
Relative to the one-stage proportional mean, Completion
reduces MSE for none of the 19 AQA systems, Composite
for one, and Pooling for 14.
The choice of estimator
changes whether the procedure improves on the baseline.
Pooling also has an estimated absolute bias of up to
1.83 percentage points, whose square is included in
the reported MSE.

We then hold Pooling fixed and compare the quota rules.
Score-only quotas with Pooling reduce MSE relative to
the proportional mean for 17 of 19 AQA systems.
They also have lower MSE than learned quotas with
Pooling for all 19 systems.
Thus, although the learned procedure with Pooling
often improves on proportional sampling, the score-only
procedure performs better in every AQA system examined
at this budget.
The original learned rule provides no additional MSE
improvement over quotas determined without using
pilot outcomes.

These results suggest two checks before deploying a
pilot-based allocation rule.
First, choose how the acquired labels will enter the
estimate: the same sampling design can lead to different
MSE and bias under different estimators.
Second, compare learned quotas with both proportional
and score-only quotas while holding that estimator
and the total label budget fixed.
When relevant historical annotations are already available, they can
support this comparison through replay, with every simulated pilot
label charged to the evaluation budget. A newly purchased validation
pool adds a separate cost and is not a prerequisite for using the
proportional stratified mean.
For the AQA point-estimation comparisons here, score-only
quotas with Pooling provide a stronger reference than
the proportional mean alone, although the proportional
mean remains better for two systems.
An adaptive rule should demonstrate an improvement over
the appropriate fixed-quota alternative.
When uncertainty intervals are required, their coverage
must also be evaluated under the chosen sampling design
and estimator.
The next subsection tests whether shrinking the learned
quotas makes them competitive with the score-only reference.

\subsection{Can shrinkage make learned quotas competitive?}
\label{sec:shrinkage}

We next test whether moderating the learned quotas improves
on score-only allocation.
For each stratum, the first rule averages the final label
counts assigned by the learned and proportional rules.
The second averages the counts assigned by the learned
and score-only rules.
Each average gives the two rules equal weight, with the
same mixing weight used for every system.
The pilot samples and total label budgets remain unchanged.
Table~\ref{tab:r40-shrinkage} holds Pooling fixed to
compare the quota rules.
Appendix~\ref{app:r40-shrinkage} gives the allocation
implementation, results for the other estimators and preference configurations, and the paired Monte Carlo comparison protocol.

\begin{table}[htbp]
\centering
\small
\setlength{\tabcolsep}{4pt}
\caption{
\textbf{Shrinkage improves MSE, but ordinary pooled intervals
can still undercover.}
All methods use Pooling on the 19 AQA systems, with matched
pilot samples and total label budgets.
The reference is score-only quotas with Pooling.
"Lower MSE" counts systems with lower MSE than this reference.
MSE ratios are summarized by their median and range across systems.
Coverage is the percentage of repeated samples whose ordinary
pooled 95\% normal interval contains the full-pool target;
the last column gives its range across systems.
The paired-comparison protocol and Monte Carlo checks are given in Appendix~\ref{app:r40-shrinkage}.
}
\label{tab:r40-shrinkage}
\begin{tabular}{@{}lccc@{}}
\toprule
Final-quota rule
& Lower MSE
& \shortstack{MSE ratio\\median (range)}
& \shortstack{95\% CI coverage\\range (\%)} \\
\midrule
Score-only (reference)
& ---
& 1.000
& 93.75--95.08 \\

Learned
& 0/19
& 1.096 (1.023--1.716)
& 75.24--91.22 \\

Half learned + proportional
& 8/19
& 1.016 (0.836--1.259)
& 92.17--94.02 \\

Half learned + score-only
& \textbf{17/19}
& \textbf{0.926} (0.866--1.061)
& 88.09--93.54 \\
\bottomrule
\end{tabular}
\end{table}

Averaging learned and score-only quotas reverses the
point-estimation result from the preceding subsection.
With Pooling, this mixture has lower MSE than score-only
quotas with Pooling for 17 of 19 AQA systems,
with a median MSE ratio of 0.926.
Averaging learned and proportional quotas improves on
the same reference for eight systems.
Thus, on AQA, a regularized learned rule can outperform the score-only
reference, and the shrinkage target matters. Across the eight
preference configurations, the score-directed mixture improves on
that reference only for Arena mean with Qwen
(Table~\ref{tab:r40-shrinkage-all}).

The MSE improvement does not by itself ensure reliable
uncertainty reporting.
For the mixture of learned and score-only quotas,
ordinary pooled 95\% intervals achieve only
88.09--93.54\% coverage, compared with
93.75--95.08\% for score-only quotas with Pooling
(Table~\ref{tab:r40-shrinkage}).
This mixture therefore merits consideration for point
estimation, while uncertainty-sensitive decisions require
an interval procedure validated under the adaptive design.
Section~\ref{sec:unbiased_no_CI} examines coverage failures in the original
learned-quota procedures to distinguish the roles of
estimator bias and uncertainty estimation.

\subsection{Does active acquisition improve on score stratification?}
\label{sec:active_comparison}

The preceding experiments learn how many labels to collect in each
stratum. We now consider published methods that select individual
items: active statistical inference (ASI; \citealp{zrnic2024active})
and RobustASI~\citep{li2025robust}.
Both use automatic scores to guide annotation and correct for unequal
sampling probabilities when estimating human quality.
RobustASI also uses a paid pilot to adjust the sampling rule.
We evaluate GLIDE's capped, tuned ASI implementation~\citep{martinon2026glide}
and the RobustASI adaptation to our finite-pool target.
Appendix~\ref{app:active} specifies the estimators and implementation.

We use two references because they answer different questions.
RobustASI's \emph{same-pilot uniform control} keeps its pilot,
estimator, and budget setting. It gives remaining items equal
sampling probabilities within each original group.
It tests whether selective acquisition helps within that procedure.
The \emph{proportional stratified human mean} instead uses the available
scores to form strata and spends the label budget within those strata.
It tests whether the complete active procedure improves on a simpler
way to use the same information.

\begin{figure}[htbp]
\centering
\includegraphics[width=0.7\linewidth]{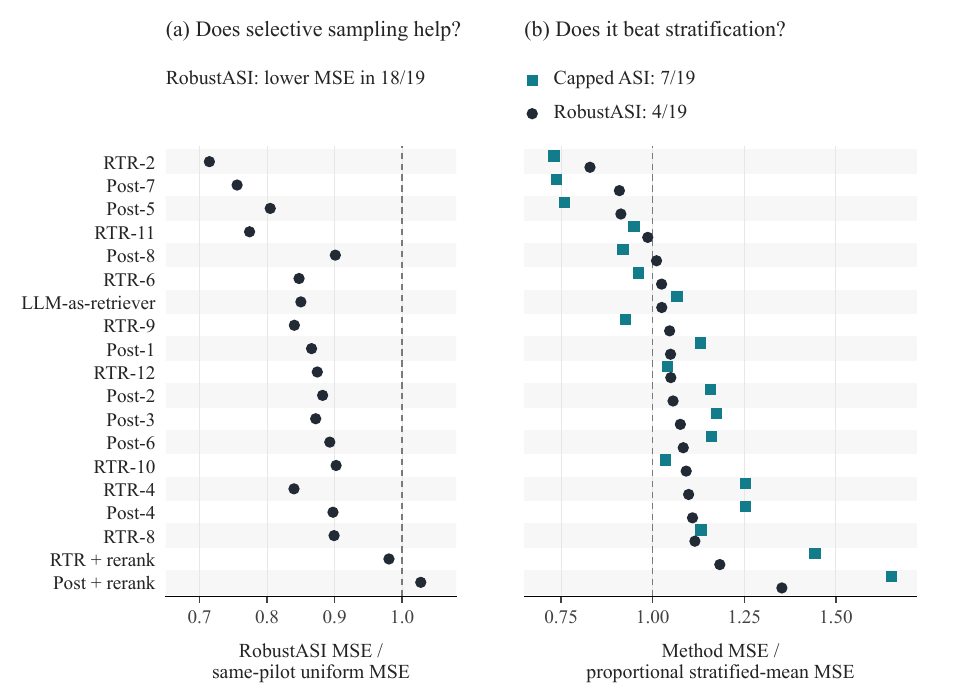}
\caption{\textbf{The reference changes the apparent benefit of active acquisition.}
AQA at nominal budget $L=250$.
(a) RobustASI relative to its same-pilot uniform control;
(b) capped ASI and RobustASI relative to the proportional
ten-stratum human mean. Ratios below one favor the active method.
Counts use point estimates; Monte Carlo standard errors are reported
in Table~\ref{tab:active_errors}.
}
\label{fig:aqa}
\end{figure}

Figure~\ref{fig:aqa} shows how this choice changes the conclusion.
At a nominal budget of 250 labels, RobustASI has lower MSE than its
uniform control for 18 of 19 AQA systems, but lower MSE than the
proportional stratified mean for four.
Capped ASI improves on the stratified mean for seven systems.
On AQA (Table~\ref{tab:active_errors}), selective acquisition usually helps within RobustASI, but the
complete procedure improves on the stratified mean for fewer systems. In the preference comparisons, its benefit over the same-pilot
uniform control is concentrated in HelpSteer3 (Table~\ref{tab:pref_active}).
Monte Carlo standard errors for these MSE ratios are also reported in these two tables.

The methods also differ in how they spend labels.
Capped ASI uses at most $L$ labels. RobustASI controls expected
expenditure, so its realized label count can exceed $L$.
This occurs in about 48\% of runs in the primary AQA experiments.
All pilot labels count toward this expenditure.
We therefore calculate a reference risk for each run using the active
method's actual label count. Specifically, we compute the exact MSE
of a proportional stratified mean with that many labels.
We then average these reference risks across runs.
The AQA counts remain seven for capped ASI and four for RobustASI.
Thus differences in label expenditure do not explain the change in
conclusion when uniform sampling is replaced by score stratification.
Appendix~\ref{app:active_results} gives these results and the preference
comparisons. Further checks in Appendix~\ref{app:r38_active} vary the
RobustASI error model and test selective acquisition within fixed
score strata.

For deployment, compare an active procedure with the proportional
stratified mean using the same scores and charging all pilot labels.
If the annotation budget is a hard limit, use a procedure designed
for that limit; an expected-budget result alone does not establish
performance under a hard cap.
This comparison tests whether the complete active procedure improves
mean estimation beyond the static score-stratified reference. Section~\ref{sec:decisions} asks whether lower MSE
also improves system selection and uncertainty reporting.

\section{Do Better Mean Estimates Support Better Decisions?}\label{sec:decisions}
So far, we have compared procedures by the MSE of a system's estimated
human quality. Evaluation teams also use these estimates to choose
between systems or to report uncertainty. These decisions require
additional checks: ranking depends on errors in the differences
between system estimates, and confidence intervals depend on how well
their uncertainty estimates describe repeated sampling.
We examine these questions using the AQA leaderboard and the pilot
estimators studied above.

\subsection{Better mean estimates need not yield better rankings}

Prediction-powered ranking uses human and model pairwise
comparisons to quantify uncertainty about model ranks
\citep{psr8_chatzi_ranking}.
Here, we ask whether improvements in estimating individual
system means also help select and rank systems on a fixed
evaluation pool.
Let $\theta_j$ be system $j$'s full-pool human mean and
$\widehat\theta_j$ its estimate from $L$ labels.
The estimated winner is
$\widehat j=\arg\max_j\widehat\theta_j$.
We measure how often this choice identifies the true best
system, how many system pairs are ordered incorrectly,
and how many of the true top three are recovered.
Each comparison uses $L$ labels per system, or $19L$ labels
in total.

We first compare proportional and working quotas using the same
score strata and stratified-mean estimator as in
Section~\ref{sec:scores}.
We coordinate sampling across systems by shuffling the question IDs once per replay and using that same shuffled list for all 19 systems. For each system, we select questions separately within its score strata:
a stratum assigned $n_h$ labels receives the first $n_h$ questions in the shuffled list that belong to that stratum. For example, with question order $(1,2,3,4)$ and one label per stratum, a system whose strata are $\{1,2\}$ and $\{3,4\}$ selects questions
1 and 3. A system whose strata are $\{1,3\}$ and $\{2,4\}$ instead selects questions 1 and 2.
Thus, systems can share labeled questions while retaining their own strata and quotas.
The random ordering gives each system a simple random sample within each stratum.
Appendix~\ref{app:ranking} also examines independent question orderings across systems.

\paragraph{A gain in individual accuracy can accompany worse selection.}
At 250 labels per system, working quotas reduce mean-estimation MSE
for 13 of the 19 systems relative to proportional quotas.
Yet the corresponding ranking comparison favors proportional
sampling: the \emph{Working mean} row in
Table~\ref{tab:ranking_r38} selects the best system in 97.21\% of
replays, compared with 99.07\% for the \emph{Proportional mean}.
Working quotas also have more inverted pairs and lower top-three
recall in this comparison.
Thus, improving the individual MSEs of most systems does not
establish an advantage for model selection.

\begin{figure}[htbp]
\centering
\includegraphics[width=0.7\linewidth]{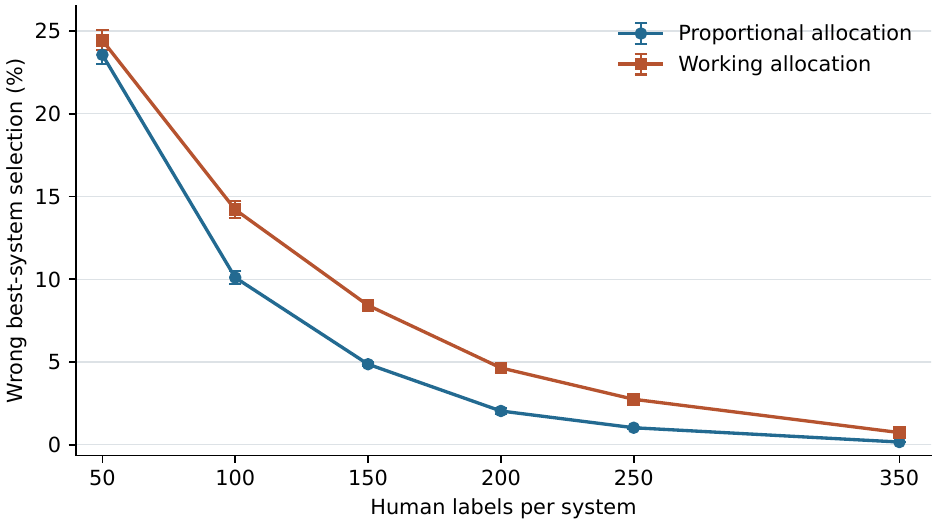}
\caption{\textbf{The label budget changes how clearly selection errors differ.}
Probability of selecting the wrong best system among 19 AQA systems.
Proportional and working quotas use the same stratified-mean
estimator and the same score strata for each system.
Each system receives exactly $L$ labels, for $19L$ per leaderboard.
Samples use the shared question ordering described in the text.
Error bars are $\pm2$ Monte Carlo standard errors.
}
\label{fig:aqa_budget}
\end{figure}

The size of the selection difference also depends on the budget.
Figure~\ref{fig:aqa_budget} varies the number of labels while keeping
the systems and their full-pool human means fixed.
At 100 labels per system, working quotas select the wrong winner in
14.21\% of replays, compared with 10.10\% for proportional quotas.
The increase is 4.11 percentage points
(paired MCSE: 0.26 percentage points).
At 350 labels, the error rates fall to 0.74\% and 0.16\%, respectively.
Both designs therefore appear highly accurate at the larger budget,
even though their selection errors differ appreciably when labels
are more limited.

\paragraph{Ranking depends on errors in differences between systems.}
For two systems $j$ and $k$, the relevant estimate is
$\widehat\theta_j-\widehat\theta_k$, whose variance is
\begin{equation}
 \Var(\widehat\theta_j-\widehat\theta_k)
 =\Var(\widehat\theta_j)+\Var(\widehat\theta_k)
  -2\Cov(\widehat\theta_j,\widehat\theta_k).
 \label{eq:ranking_covariance_main}
\end{equation}
When the two estimation errors move together, part of their variation
cancels in the difference.
Changing quotas can change this covariance as well as the individual
variances.
At 250 labels with shared question ordering, the reduction in
positive covariance accounts algebraically for about 80\% of the
increase in average pairwise error variance under
working quotas. For comparisons between the true winner and its two
closest competitors, however, increases in individual variances
account for most of the increase.
At the same budget under independent
sampling, working quotas slightly improve average pairwise ordering but still reduce best-system selection accuracy
(Appendix~\ref{app:ranking}).
Thus both the ranking objective and
the relationship between system samples matter. Pairwise variance
helps explain these results, while selection probabilities also
depend on the true quality gaps and error distributions.

\paragraph{Different ranking objectives can favor different procedures.}
Table~\ref{tab:ranking_r38} also compares the pilot-based procedures
at the same total budget of 250 labels per system.
Score-only quotas with Pooling reduce mean-estimation MSE for 17 of
the 19 systems relative to the proportional mean.
They also reduce inverted pairs from 6.68\% to 6.41\%, but
best-system selection falls from 99.07\% to 98.67\%.
Here the same procedure improves average pairwise ordering while
making the choice of a single winner less reliable.
Shrinking learned quotas halfway toward score-only quotas, then
using Pooling, improves both of these ranking metrics relative to
score-only quotas with Pooling, although top-three recall is
slightly lower.
This gives a positive result for regularized adaptation, with a
tradeoff between ranking objectives.

\begin{table}[t]
\centering\small\setlength{\tabcolsep}{5pt}
\caption{AQA ranking at 250 labels per system.
Entries are percentages, with MCSE in parentheses.
Lower inverted-pair rates and higher selection and recall rates are better.
All rows use the same joint replay with shared question orderings.
The last two rows shrink learned final quotas halfway toward proportional or score-only quotas, respectively, and use Pooling.}
\label{tab:ranking_r38}
\begin{tabular}{lrrr}\toprule
Procedure & Inverted pairs & Best-system selection & Top-three recall\\\midrule
Proportional mean & 6.68 (0.01) & 99.07 (0.04) & 96.16 (0.05) \\
Working mean & 6.85 (0.01) & 97.21 (0.08) & 94.49 (0.06) \\
\addlinespace
Score-only + Pooling & 6.41 (0.02) & 98.67 (0.07) & 96.08 (0.09) \\
Learned + Pooling & 6.43 (0.02) & 99.30 (0.05) & 94.21 (0.12) \\
Proportional-shrink + Pooling & 6.27 (0.02) & 99.29 (0.05) & 96.25 (0.08) \\
Score-shrink + Pooling & 5.99 (0.01) & 99.40 (0.04) & 95.86 (0.09) \\
\bottomrule\end{tabular}
\end{table}

\paragraph{Deployment implication.}
Evaluate a sampling procedure against the decision it must support.
For choosing one model, compare best-system selection; for a shortlist,
compare top-three or top-$k$ recall; for a leaderboard, compare ordering errors.
The AQA results show why these checks matter: a procedure can improve
individual MSEs or average pairwise ordering while making winner
selection worse.

\subsection{Unbiased estimation does not ensure reliable intervals}
\label{sec:unbiased_no_CI}
A second deployment goal is to report uncertainty about a system's
quality, for example when deciding whether it meets a required threshold.
In this setting, an accurate point estimate must be accompanied by an
interval with the claimed coverage.
For each estimator $\widehat\theta$, we examine the normal interval
\begin{equation}
 I=\left[
 \widehat\theta-z_{1-\alpha/2}\sqrt{\widehat V},
 \;\widehat\theta+z_{1-\alpha/2}\sqrt{\widehat V}
 \right],
 \qquad
 \text{coverage}=\Pr(\theta\in I).
 \label{eq:normal_main}
\end{equation}
Here $\widehat V$ estimates the estimator's sampling variance, and
$z_{1-\alpha/2}=\Phi^{-1}(1-\alpha/2)$, where $\Phi$ is the standard
normal cumulative distribution function.
We set $\alpha=0.05$ and use
$z_{0.975}$ in computation.
The nominal coverage is therefore 95\% over repeated label acquisition from the fixed pool.
Static corrections use the variance estimate in
Equation~\eqref{eq:plugin}, and pilot procedures use those in Appendix~\ref{app:r38_pilot}.

\paragraph{First check the interval's width.}
We begin with learned quotas and Composite at 250 labels per AQA
system.
Both its point estimate and its variance estimate are unbiased, so
this comparison tests whether those properties are sufficient for
reliable normal intervals at the available sample size.
The observed coverage is only 84.97--92.99\% across the 19 systems.

We next ask how much of this undercoverage comes from estimating
the standard error with limited labels.
To assess this, we recompute the intervals around exactly the same
point estimates, replacing each sample-based variance estimate with
a fixed reference variance for that system and procedure.
This reference estimates the variance $V_*$ of the estimator over repeated pilot and final-stage sampling, using the complete human-labeled pool.

\begin{figure}[htbp]
\centering
\includegraphics[width=0.75\linewidth]{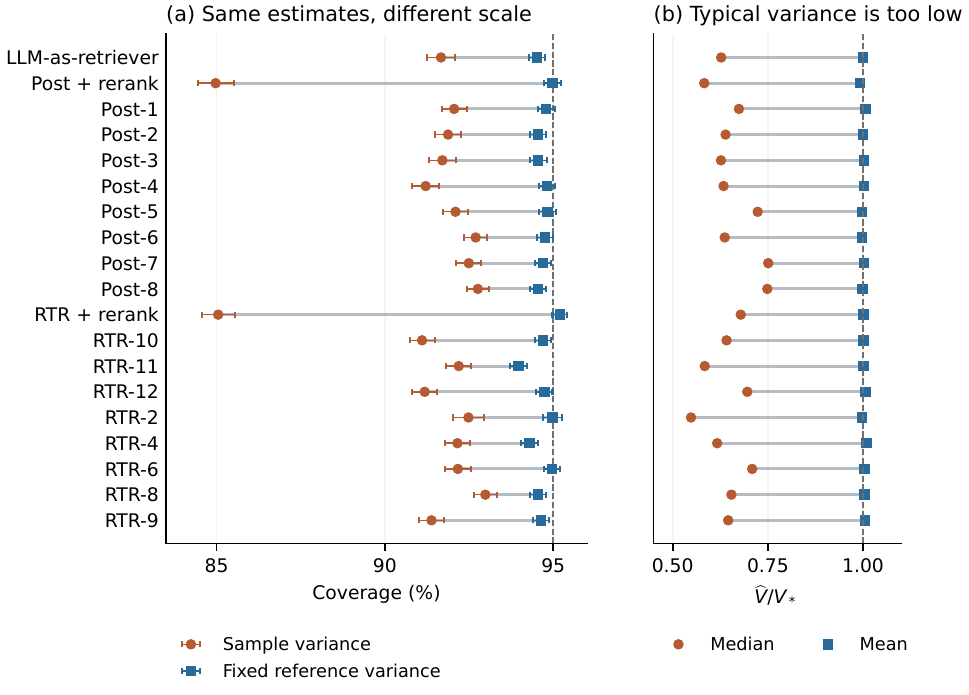}
\caption{\textbf{An accurate average variance can hide unreliable interval widths.}
Learned quotas with Composite on the 19 AQA systems.
Left: coverage of the same point estimates using either sample-based
variance or a fixed reference variance; bars are
$\pm1.96$ MCSE.
Right: the mean and median sample variance relative to the reference.
The reference uses complete pool outcomes to diagnose the source
of undercoverage.}
\label{fig:coverage_mechanism}
\end{figure}

Coverage rises from 84.97--92.99\% to 93.97--95.20\%
(Figure~\ref{fig:coverage_mechanism}).
Because the point estimates are unchanged, this improvement shows
that the estimated interval widths account for a substantial part
of the undercoverage.
The reference uses information unavailable during deployment;
its purpose is to diagnose the failure.
Appendix~\ref{app:r40-coverage} gives its calculation.

The distribution of the variance estimates further explains this difference.
Their means are 0.993--1.009 times the reference variance, but their
medians are only 0.548--0.751 times the reference.
Depending on the system, 27--46\% of replays estimate less than half
the reference variance.
Occasional large values compensate for frequent small values,
making the estimate accurate on average while often producing
an interval that is too narrow.
Unbiasedness of $\widehat V$ therefore does not ensure that the
random standard error $\sqrt{\widehat V}$ gives reliable coverage.

\paragraph{Then check the interval's center.}
Pooling uses the same acquired labels as Composite but combines them
differently and can introduce bias under learned quotas.
Its ordinary normal intervals cover only 75.24--91.22\% across AQA.
Replacing its sample-based variance with its own fixed reference
raises coverage to 86.76--95.06\%.
Keeping that fixed width and additionally subtracting the measured
bias from each point estimate raises coverage to 94.74--95.24\%.
These successive comparisons distinguish two problems: an unstable
estimate of uncertainty and a systematic displacement of the estimate.
Appendix~\ref{app:r40-coverage} gives the protocol and uncertainties.
The shrinkage results in Section~\ref{sec:shrinkage} show that reducing
Pooling's bias can improve MSE while its ordinary intervals still undercover.

\paragraph{Limited labels make interval validation necessary even without adaptation.}
At 100 labels in HelpSteer3, ten bins per group leave 30 of 40 strata
with only two observations, and a two-wave design requiring two
observations per stratum in each wave is infeasible.
In AQA, the static working mean for Post + reranking has only 85.53\%
normal coverage at the same per-system budget.
Two observations suffice to calculate a sample variance, but do not
ensure a stable standard error.
Table~\ref{tab:calibration_coverage} reports further static correction
results for the configurations in Table~\ref{tab:nonlinear_r38};
several also fall below nominal coverage.

\paragraph{Deployment implication.}
Compare interval widths only after checking coverage under the
intended sampling and label-reuse procedure.
Existing fully labeled validation pools allow this check through
repeated acquisition at the planned budget.
If coverage is inadequate, inspect estimation bias and variance
estimation separately.
Merging sparsely sampled strata or assigning them more labels are
candidate changes whose coverage must then be rechecked.
The evidence supports evaluating the complete estimator--interval
pair: a lower MSE or an unbiased variance estimate alone does not
justify a 95\% confidence claim.

\section{Conclusions and Guidance for Human Evaluation}
\label{sec:revision_conclusion}

The main lesson is that the value of an automatic score depends on how
the evaluation procedure already uses it. Score stratification can
capture information that would otherwise benefit a fitted correction.
Allocating more labels to apparently difficult responses can help, but
only when the allocation proxy identifies variation in human outcomes.
Learning that allocation introduces both estimation noise and a choice
about how to reuse the pilot. These interactions explain why a complete
procedure can outperform a score-unaware reference while contributing
little beyond a simpler procedure using the same scores.

Our contribution is to make those incremental comparisons explicit and
connect them to their statistical causes and practical consequences.
The evidence includes useful gains from unequal allocation in several
pools and from regularized pilot adaptation on AQA, alongside settings
where these additional steps do not improve the chosen reference. It also shows
why their value must be judged against the output being purchased:
an accurate mean, a correct system choice, or a reliable uncertainty
statement. Lower individual MSE alone does not establish improvements
in the latter two. This leads to different guidance for deploying an
evaluation procedure and for testing a new one.

\paragraph{Choosing a procedure when labels are costly.}
If a team has scores but no relevant historical human labels, our
recommended starting procedure is proportional sampling within score
strata followed by the stratified human mean. Fix the partition and
quotas before annotation, sample uniformly within each stratum, and use
the target's population weights when combining the human averages.
Choose the number of strata to suit the budget and natural groups.
Five to ten bins are candidates supported by our experiments; smaller
budgets may require fewer. This starting procedure
avoids spending a separate pilot merely to decide how to spend the remaining labels.

When relevant historical annotations already exist, use them to compare
a small set of alternatives before collecting the current evaluation
labels. Match the judge, annotation protocol, target, and intended
budget as closely as possible. Compare a regularized correction on the
same samples, and compare score-only or learned allocations with the
estimator held fixed. Choose and tune the procedure separately from
the data used to assess its performance. Prefer the simpler procedure
when an advantage is too small or uncertain to justify another fitting
step or annotation round. Without such prior evidence, the present
results do not provide a score-only rule that can reliably select the
best procedure for a new pool. Newly purchased validation labels should belong
in the cost calculation. 

Our AQA results show that an allocation can lower individual-system
MSE while increasing the probability of selecting the wrong best
system. Ranking performance also depends on how samples are selected
across systems: the pairwise-ordering comparison between working and
proportional quotas changes between shared-priority and independent
sampling. For best-system selection, therefore, compare allocations
by their selection error at the intended label budget, using the
planned rule for selecting human-labeled samples across systems.

If the evaluation must report confidence intervals, compare coverage
and width alongside point-estimation MSE.
In our learned-quota experiments, ordinary pooled intervals
undercover even when Pooling gives favorable MSE.
These intervals should not be reused unchanged after pilot outcomes
determine the final quotas.
Fixed sampling avoids this outcome-dependent pooling problem,
although small strata can still make normal intervals unreliable.

The spending limit places a separate constraint on acquisition.
If at most $L$ labels can be purchased, choose fixed quotas or a
procedure designed to enforce that cap.
Native RobustASI \citep{li2025robust} controls expected expenditure and exceeds the
nominal AQA budget in about 48\% of replays.
Stopping it at $L$ labels changes the acquisition rule, so the
estimator must be reassessed under that modification.
Count every newly acquired system--question judgment, including
pilot labels, even when questions are shared across systems.
If automatic scores must also be purchased, deduct their cost before
setting the human-label budget.

\paragraph{Evaluating a proposed improvement.}
Researchers with fully annotated pools can identify where a
procedure's gains come from and whether they improve the intended
evaluation output. We recommend four comparisons:
\begin{enumerate}
\item Fix the target, annotation unit, available scores, and total
budget. Count every pilot label, distinguish exact, capped, and
expected expenditure, and keep unacquired outcomes hidden from
the procedure.

\item Compare the original-group mean with the proportional
score-stratified mean to measure the gain from stratification.
Then compare corrections on common samples and allocations under
a common estimator to measure their additional contributions.

\item For pilot methods, cross fixed and learned quota rules with
the label-reuse estimators. Include fixed score-only quotas at
the same total budget.
For active methods, report both their matched uniform control
and the score-stratified reference, together with actual expenditure.

\item Compare the required output across relevant budgets.
Report paired Monte Carlo uncertainty for MSE and ranking differences,
and interval width alongside coverage.
For rankings, reproduce the intended sharing of questions across
systems. Keep method selection and tuning separate from the
replays used for the final comparison.
\end{enumerate}

These recommendations concern the fixed pools, judges, and annotation
protocols studied here.
Further work is needed to choose procedures
reliably from limited historical labels, test their transfer to new
pools, and evaluate alternative intervals for the adaptive procedures
that perform well here. 
The practical starting point is to use scores to organize a simple
human sample. Further correction or adaptation should earn its place
by improving the required output at the full evaluation cost.
\label{main:end}
\FloatBarrier
\clearpage
\bibliographystyle{references_style}
\bibliography{references}
\clearpage
\appendix
\section{Data and Experimental Protocol}\label{app:protocol}
\subsection{Targets and annotation units}
AlpacaFarm~\citep{dubois2023alpacafarm} uses all 648 released human-evaluation pairs. The outcome is
the fraction of votes favoring response B. The 2,596 votes comprise
645 pairs with four votes, two with five, and one with six; acquiring a
label reveals the released aggregate. Arena~\citep{chiang2024arena} uses 2,870 battles with
exactly one GPT-4-family response, encoding its win, tie, or loss as
$1,1/2,0$. Claude, GPT-3.5, and other opponent families define known
groups. The contrast subtracts the observed preference means against
Claude and GPT-3.5 on their 1,247 combined items; their prompt populations
need not coincide.

HelpSteer3~\citep{wang2025helpsteer3} uses all 2,017 validation pairs in code, general, multilingual,
and STEM groups. Preference for response 1 uses $1,1/2,0$. The release
contains 5,950 retained individual annotations. High-disagreement
filtering occurred before release, so neither the unfiltered population
nor the labor spent on excluded annotations is recoverable.

AttributedQA~\citep{bohnet2022attributed} supplies automatic attribution scores and aggregated
human labels for 19 systems evaluated on the same 1,000 questions. Human attribution decisions use five-rater majorities
\citep{bohnet2022attributed}.
Repeated outputs across systems may share annotations, so multiplying
system--question labels by five does not recover the number of unique
human judgments.

\subsection{Scores and information access}
The preference score is the designated response's win
probability plus half its tie probability.
Both judges score each pair in both response orders;
reversed probabilities are reoriented before averaging.
Mistral scores come from
\texttt{Mistral-Small-3.2-24B-Instruct-2506}
\citep{psr8_mistral_small32}, using normalized next-token
likelihoods with BF16 weights.
Qwen scores come from \texttt{Qwen3.8-27B}
\citep{qwen2026model}, using Unsloth's
\texttt{UD-IQ4\_XS} GGUF quantization.\footnote{Quantized
weights: \url{https://huggingface.co/unsloth/Qwen3.8-27B-GGUF}.}
This model generates the probability vectors as JSON
and is served under the identifier
\texttt{qwen3.8:27b-ud-iq4-xs}.
Judge prompts exclude model identities and human outcomes.

Procedures receive scores, target weights, permitted grouping metadata,
and acquired human labels. Complete outcomes define the target and
support exact-risk calculations and retrospective diagnostics. Masking
tests verify that unacquired outcomes cannot change an estimate.
Compatible designs share random priorities, and the ranking protocol
specifies how priorities are shared across systems. 

\subsection{Budgets and precision}
Static and strict-pilot designs use integer budgets and require at least two labels per sampled stratum in each wave. Infeasible designs are not run. Minimum counts and rounding determine the realized budget and pilot fractions.

The primary correction comparisons use 10,000 paired draws and
100 independent batches for Monte Carlo standard errors (MCSEs).
Nonlinear calibration uses 5,000 draws.
Pilot and shrinkage comparisons use 2,000 independent pilots.
We integrate final-stage MSE analytically and use 20 final draws per
pilot to evaluate coverage and ranking metrics.
The static and capped-ASI arms of the primary AQA and preference
active-method comparisons use 100,000 draws.
RobustASI uses 200 pilots with 100 final draws each
\citep{zrnic2024active,li2025robust}.
The ranking and small-budget analyses give their own draw counts and
MCSE calculation units below.

\section{Sampling Details and Risk Derivations}\label{app:methods}
\subsection{Design assumptions and variance estimation}\label{app:estimators}
All outcomes and scores are fixed.
Samples are drawn without replacement and independently
across strata, conditional on fixed counts.
Target weights are constant within each stratum.
The Arena contrast assigns item weights $N/|G_+|$ to items
in $G_+$ and $-N/|G_-|$ to items in $G_-$, so its target
is the difference between the two observed-pool means.
We report risk relative to the proportional stratified-mean reference, which is $R/R_0$.

For fixed coefficients in Equation~\eqref{eq:fixed}, the full-score term is
constant and the sampled residual mean is unbiased for the stratum
residual mean. Its SRS variance is
$(1-n_h/N_h)S^2_{y-b_hs,h}/n_h$, where $S^2$ uses divisor $N_h-1$.
Independence across strata gives Equation~\eqref{eq:fixedrisk}
\citep{sarndal1992model,cochran1977sampling}.
The fitted coefficients in Equation~\eqref{eq:coefficients} instead use
sample covariance with divisor $n_h-1$ and known full-stratum score
variance with divisor $N_h-1$. The local rule matches the local point estimator implemented in GLIDE~\citep{martinon2026glide}. The fixed-coefficient calculation leaves out uncertainty
from fitting. Accounting for fitted models is central to
model-assisted estimation \citep{sarndal1992model} and
semiparametric debiasing
\citep{vanderlaan2006targeted,chernozhukov2018double,chen2026kernel,liu2026one}.
For the unrestricted local and group-shared fits, coefficients are
estimated on the same labels used in the final estimate.
We evaluate their MSE by replay, including any fitting bias.

For the normal interval in Equation~\eqref{eq:normal_main}, the static
residual variance estimate is
\begin{equation}
 \widehat V=\sum_h c_h\widehat\Var_{S_h}(y-\widehat b_hs).
 \label{eq:plugin}
\end{equation}
Here, $\widehat{\operatorname{Var}}_{S_h}$ is the sample
variance of the residuals in $S_h$, with divisor $n_h-1$. It is unbiased for fixed coefficients and a residual linearization for
fitted coefficients. We use $z_{0.975}=\Phi^{-1}(0.975)$, the standard-normal critical
value defined in the main text.

\subsection{Exact quotas and the constrained oracle}\label{app:allocation}
Let $B_g$ be the number of score bins within natural group $g$.
Stable score sorting defines bin boundaries at
$\lfloor kN_g/B_g\rfloor$, for $k=0,\ldots,B_g$.
Stable ranks can split ties.
The proportional rule uses importance $u_h=|\omega_h|$.
The working rule uses $u_h=|\omega_h|\sqrt{v_h^{\rm work}}$,
with the annotation-specific variance defined in the main text and
derived in Appendix~\ref{app:r45_variance} \citep{fisch2024stratppi}. The same variance model is used for static
working quotas and score-only final quotas. Human outcomes do not
enter this proxy.

We find a scale $t$ by 100 bisection steps such that
$\sum_h\min\{N_h,\max\{2,tN_hu_h\}\}=L$.
Counts are floored, then remaining labels are assigned by decreasing
fractional remainder with stable tie breaking. This enforces the
minimum, capacity, and exact budget. Individual zero importances are replaced by $10^{-12}$.
An all-zero importance vector is replaced by ones, giving size-proportional
counts subject to the same bounds.
For score-only and learned final quotas, the routine uses remaining sizes
$M_h$ and budget $L-P$.
The matched proportional control is set separately to
$q_h^p=n_h^0-m_h$, so its combined pilot and final counts recover the one-stage proportional allocation exactly.

The full-label oracle minimizes Equation~\eqref{eq:fixedrisk} subject to
the same integer constraints. Its finite-population term is constant
in the counts. For the mean-estimator allocation oracle, we set $b_h=0$,
so $S^2_{e,h}=S^2_{y,h}$ in this calculation. An additional label in stratum $h$ reduces risk by
\begin{equation}
 d_h(n_h)=\frac{a_h^2S^2_{e,h}}{n_h(n_h+1)}.
\end{equation}
Starting at the lower bounds, repeatedly choose the largest feasible
reduction. These reductions are nonincreasing within each stratum, so
an exchange argument gives an integer optimum
\citep{wright2014allocation}. Small-problem enumeration and feasible
exchange checks verify the implementation. This oracle requires the
complete outcomes and is used only as a diagnostic.
At the primary budget, its MSE reduction relative to proportional quotas
is 0.97--3.33\% across the six Alpaca/Arena configurations, 7.54\% and
11.09\% for HelpSteer3, and 7.80--69.34\% across AQA.

\section{Correction Controls and Interval Coverage}
\label{app:r38_correction}
\subsection{Coefficient fitting and regularization}
\label{app:revision_fitting}
The linear fits use the known full-stratum score variance in their
denominator. Local fits estimate one coefficient per stratum;
group-shared fits estimate one coefficient across the strata in an
original group. Clipping replaces either fitted coefficient by
$\min\{1,\max\{0,\widehat b\}\}$.

Following \citet{eyre2025regression}, Ridge-PPI instead uses sample
score moments and an unpenalized intercept in each stratum. The local
coefficient is
$\widehat C_{ys,h}/(\widehat S^2_{s,h}+\alpha)$;
the group-shared coefficient is defined in the main text.
Sample moments use divisor $n_h-1$.
The reported positive-penalty fits select
$\alpha\in\{10^{-6},10^{-5},\ldots,10^2\}$ by leave-one-out prediction
error on the acquired labels. Observation weights $c_h/(n_h-1)$ remain
fixed during refits. Errors within $10^{-12}$ of the minimum favor the
larger penalty, and the selected coefficient is then refitted on all
acquired labels. With two observations in a local fit, all penalties
tie and the rule selects $10^2$. These choices specify our finite-pool
adaptation. Its confidence intervals use the residual variance estimate in
Equation~(\ref{eq:plugin}).

\paragraph{Matched results for sharing and clipping.}
Table~\ref{tab:coefficient_controls} compares the local and group-shared
linear fits described in the main text, together with their clipped versions.
All four rules use the known full-stratum score variances.
They share the same partition, proportional quotas, and sampled labels,
so the comparison isolates coefficient sharing and clipping.

On the preference configurations, sharing reduces the excess MSE of
unrestricted local fits, while clipping brings both versions close to
the stratified-mean reference. Several AQA systems retain modest gains
from correction. Stabilizing a coefficient can therefore improve a
fitted estimator even when its additional benefit over the stratified
human mean is small.

\begin{table}[htbp]
\centering\small
\setlength{\tabcolsep}{3pt}
\caption{Coefficient sharing and clipping at the primary budgets,
using ten score bins per natural group and proportional quotas.
Entries are MSE divided by $R_0$, the risk of the matching stratified
human mean; lower is better. Parentheses give one Monte Carlo standard
error from 10,000 paired draws. Local fits estimate one coefficient per
stratum; Shared fits estimate one coefficient per natural group.
M/Q denote Mistral/Qwen; AQA uses AutoAIS.
The Shared column reproduces the proportional-allocation Linear
results in the main table.}
\label{tab:coefficient_controls}
\begin{tabular}{llrrrr}
\toprule
& & \multicolumn{2}{c}{Unrestricted}
  & \multicolumn{2}{c}{Clipped to $[0,1]$} \\
\cmidrule(lr){3-4}\cmidrule(lr){5-6}
Target / system & Score & Local & Shared & Local & Shared \\
\midrule
\multicolumn{6}{l}{\textit{A. Preference targets}} \\
Alpaca & M & 1.076 (0.0052) & 1.009 (0.0014) & 1.002 (0.0009) & 1.002 (0.0009) \\
Alpaca & Q & 1.032 (0.0053) & 1.005 (0.0022) & 0.994 (0.0016) & 1.000 (0.0017) \\
Arena mean & M & 1.062 (0.0045) & 1.008 (0.0018) & 1.002 (0.0010) & 1.002 (0.0010) \\
Arena mean & Q & 1.041 (0.0045) & 1.002 (0.0016) & 0.998 (0.0014) & 0.998 (0.0013) \\
Arena contrast & M & 1.067 (0.0053) & 1.008 (0.0022) & 1.000 (0.0011) & 0.999 (0.0012) \\
Arena contrast & Q & 1.079 (0.0057) & 1.010 (0.0017) & 1.003 (0.0012) & 1.002 (0.0012) \\
HelpSteer3 & M & 1.066 (0.0060) & 1.002 (0.0025) & 0.995 (0.0012) & 0.995 (0.0013) \\
HelpSteer3 & Q & 1.089 (0.0059) & 1.006 (0.0025) & 1.001 (0.0014) & 0.997 (0.0015) \\
\midrule
\multicolumn{6}{l}{\textit{B. AttributedQA: all complete systems}} \\
LLM-as-retriever & AutoAIS & 1.022 (0.0042) & 0.990 (0.0031) & 0.988 (0.0021) & 0.984 (0.0024) \\
Post + rerank & AutoAIS & 1.018 (0.0050) & 1.003 (0.0022) & 1.001 (0.0016) & 0.999 (0.0018) \\
Post-1 & AutoAIS & 1.041 (0.0046) & 1.001 (0.0026) & 1.002 (0.0023) & 0.998 (0.0022) \\
Post-2 & AutoAIS & 1.035 (0.0040) & 1.002 (0.0028) & 0.996 (0.0020) & 0.996 (0.0022) \\
Post-3 & AutoAIS & 1.015 (0.0055) & 1.000 (0.0028) & 0.985 (0.0023) & 0.993 (0.0022) \\
Post-4 & AutoAIS & 1.039 (0.0041) & 1.006 (0.0021) & 0.998 (0.0018) & 1.001 (0.0016) \\
Post-5 & AutoAIS & 1.010 (0.0050) & 0.978 (0.0034) & 0.979 (0.0025) & 0.974 (0.0026) \\
Post-6 & AutoAIS & 1.035 (0.0042) & 1.009 (0.0017) & 1.001 (0.0016) & 1.004 (0.0014) \\
Post-7 & AutoAIS & 1.027 (0.0051) & 1.010 (0.0019) & 1.007 (0.0018) & 1.004 (0.0016) \\
Post-8 & AutoAIS & 0.998 (0.0051) & 0.969 (0.0039) & 0.967 (0.0023) & 0.965 (0.0025) \\
RTR + rerank & AutoAIS & 1.007 (0.0058) & 1.003 (0.0020) & 0.994 (0.0019) & 1.000 (0.0018) \\
RTR-10 & AutoAIS & 1.026 (0.0051) & 1.005 (0.0019) & 0.997 (0.0015) & 1.001 (0.0016) \\
RTR-11 & AutoAIS & 1.010 (0.0047) & 0.991 (0.0027) & 0.988 (0.0023) & 0.984 (0.0023) \\
RTR-12 & AutoAIS & 1.009 (0.0055) & 0.988 (0.0029) & 0.986 (0.0019) & 0.984 (0.0020) \\
RTR-2 & AutoAIS & 1.000 (0.0057) & 0.988 (0.0036) & 0.978 (0.0035) & 0.982 (0.0033) \\
RTR-4 & AutoAIS & 1.005 (0.0051) & 0.977 (0.0035) & 0.978 (0.0024) & 0.973 (0.0026) \\
RTR-6 & AutoAIS & 1.007 (0.0052) & 0.980 (0.0034) & 0.983 (0.0021) & 0.977 (0.0022) \\
RTR-8 & AutoAIS & 1.009 (0.0051) & 0.996 (0.0030) & 0.983 (0.0019) & 0.990 (0.0022) \\
RTR-9 & AutoAIS & 1.012 (0.0054) & 1.002 (0.0022) & 0.989 (0.0020) & 0.998 (0.0018) \\
\bottomrule
\end{tabular}
\end{table}

\subsection{Nonlinear calibration implementation}\label{app:sensitivity}
The controls in Table~\ref{tab:nonlinear_r38} use the official
\texttt{ppi\_aipw} implementation of Calibeating
PPI~\citep{vanderlaan2026calibeating}.
LinearCal, MonoSpline, and IsoCal keep their upstream defaults:
MonoSpline has up to six internal knots and its fixed penalty; IsoCal
uses the default one-round XGBoost backend \citep{chen2016xgboost}. Each calibrator pools acquired
labels within an original target group using inverse-inclusion weights.
Signed target weights enter aggregation, not training. Every calibration
label counts in $L$; no AutoCal selection is performed.
Equation~\eqref{eq:nonlinear_main} gives the adapted point estimate.
For its interval, we use the variance construction in
Equation~\eqref{eq:plugin}, replacing the
linear residual by $y_i-\widehat f_g(s_i)$ for $i\in G_g$.

\paragraph{Coverage of the calibration estimators.}
Table~\ref{tab:calibration_coverage} reports coverage for the same
methods, partitions, budgets, and eight preference configurations
as Table~\ref{tab:nonlinear_r38}. It uses the residual normal
intervals associated with these finite-pool estimators.

\begin{table}[htbp]
\centering
\small
\setlength{\tabcolsep}{8pt}
\renewcommand{\arraystretch}{1.08}
\caption{
Coverage of nominal 95\% residual normal intervals for the
calibration comparison. Each entry gives the range of empirical
coverage across the eight preference configurations at the primary
budget. The corresponding point-estimation results appear in
Table~\ref{tab:nonlinear_r38}.
}
\label{tab:calibration_coverage}
\begin{tabular}{llr}
\toprule
Bins/group & Calibration & Coverage (\%) \\
\midrule
1 & IsoCal           & 92.42--94.26 \\
  & LinearCal        & 94.52--95.28 \\
  & MonoSpline       & 93.92--94.80 \\
  & Local correction & 94.50--95.42 \\
\midrule
5 & IsoCal           & 92.98--94.60 \\
  & LinearCal        & 94.46--95.52 \\
  & MonoSpline       & 94.20--95.24 \\
  & Local correction & 93.70--94.90 \\
\midrule
10 & IsoCal           & 93.00--94.74 \\
   & LinearCal        & 94.50--95.28 \\
   & MonoSpline       & 94.44--95.10 \\
   & Local correction & 92.42--93.68 \\
\bottomrule
\end{tabular}
\end{table}

\section{Allocation Diagnostics and Controls}\label{app:r38_mechanism}
\subsection{Working variances for the released annotation protocols}
\label{app:r45_variance}

Equation~\eqref{eq:working} applies the law of total variance to an item
chosen uniformly within a stratum and an outcome drawn from its working
distribution \citep{fisch2024stratppi}. For the preference
datasets, we first average each judge's response-order probability vectors
after orienting both toward the designated response. Their moments provide
the unchanged scalar scores and the annotation-specific variances below. The evaluation
target remains the fixed pool of released human outcomes; only the
allocation proxy uses this working distribution.

\paragraph{AQA: a binary attribution label.}
The working Bernoulli variance is $u_i=s_i(1-s_i)$. Adding the
within-stratum variation of $s_i$ gives
$v_h^{\rm work}=\overline s_{U_h}(1-\overline s_{U_h})$.

\paragraph{Arena and HelpSteer3: win, tie, or loss.}
For values $(1,1/2,0)$ and probabilities $(p_{W,i},p_{T,i},p_{L,i})$,
\[
s_i=p_{W,i}+\tfrac12p_{T,i},
\qquad
u_i=p_{W,i}+\tfrac14p_{T,i}-s_i^2.
\]
The resulting stratum variance is
\[
v_h^{\rm work}
=\overline s_{U_h}(1-\overline s_{U_h})
 -\tfrac14\overline p_{T,U_h}.
\]
Here,
$\overline p_{T,U_h}=N_h^{-1}\sum_{i\in U_h}p_{T,i}$
is the mean predicted tie probability in stratum $h$.
Tie probability therefore contributes explicitly to the
uncertainty of the numerical human outcome.

\paragraph{Alpaca: the average of several binary votes.}
Alpaca's released $y_i$ averages $r_i$ annotator votes
\citep{dubois2023alpacafarm}. The pool contains 645 items with four votes,
two with five, and one with six, totaling 648 items and 2,596 votes.
There are no individual tie votes; $y_i=1/2$ represents equally many
votes for each response.
Assigning each vote working mean $s_i$ and assuming conditional
independence gives
\[
u_i=\frac{s_i(1-s_i)}{r_i},
\qquad
v_h^{\rm work}
=\frac{1}{N_h}\sum_{i\in U_h}\frac{s_i(1-s_i)}{r_i}
 +\frac{1}{N_h}\sum_{i\in U_h}(s_i-\overline s_{U_h})^2.
\]
The vote counts are known acquisition-protocol metadata. Human vote
preferences do not enter the quotas. Only the within-item component
is divided by $r_i$; variation in predicted means across items remains.
If all items had four votes, the expression would be
$\tfrac14\overline s_{U_h}(1-\overline s_{U_h})+
\tfrac34\operatorname{Var}_{U_h}(s)$, with population divisor $N_h$.

Conditional independence is a working assumption. To check dependence,
we also use a common pairwise within-item vote correlation
$\rho\in\{0.25,0.5\}$, replacing the item variance by
\[
u_i(\rho)=s_i(1-s_i)\left\{\rho+\frac{1-\rho}{r_i}\right\}.
\]
These fixed sensitivities give the same primary integer quotas as
$\rho=0$ for both judges. At ten bins and $L=162$, mean-MSE ratios
remain $0.99619$ for Mistral and $1.02788$ for Qwen.
We fix $\rho$ at these values to assess sensitivity to within-item dependence.

\subsection{Confidence rescaling with fixed strata}
\label{app:r45_temperature}

Model reliability depends on both probability calibration
\citep{guo2017calibration} and how errors are distributed
across inputs. Group-level and example-level diagnostics
can reveal failures hidden by aggregate performance
\citep{sagawa2020distributionally,swayamdipta2020cartography,
xue-etal-2026-supervised}.
Studies of judge biases and black-box auditing provide
further reasons to examine these error patterns
\citep{zheng2023judging,liu2026explanations}.
For allocation, the relevant question is whether a better
calibrated score also directs labels toward strata with
greater variation in human outcomes. We apply the temperature transformation from the main text to the
binary outcome probabilities for AQA and individual Alpaca votes, and
to the win/tie/loss probabilities for Arena and HelpSteer3
\citep{guo2017calibration}.
The transformed probabilities determine the working mean and variance;
Alpaca retains its actual vote counts and conditional-independence
model. Strata, outcomes, budgets, and the mean estimator remain fixed,
so the intervention changes acquisition alone. For a three-category
prediction, sharpening can favor a tie and need not move its numerical
mean away from $1/2$.

We measure mean calibration and quota distance by
\[
 E_{\rm cal}(T)=\sum_h\frac{N_h}{N}
       \left|\overline s_{U_h,T}-\overline y_{U_h}\right|,
 \qquad
 D_{\rm quota}(T)=\frac{1}{2L}
       \sum_h\left|n_h(T)-n_h^{\rm prop}\right|.
\]
Calibration uses positive size weights even for the signed Arena
contrast. Quota distance is the fraction of label slots that must move
between strata to recover proportional counts. A moved slot removes
both an excess and a deficit, giving the factor $1/2$.

\begin{table}[t]
\centering
\caption{Temperature sensitivity across the 27 primary configurations.
Entries count decreases relative to $T=1$. Configurations share data
and are descriptive comparisons, not independent replications.}
\label{r6:tab:temperature}
\small
\begin{tabular}{lrrr}
\toprule
Temperature & Calibration error & Quota distance & MSE \\
\midrule
$T=0.5$ & 3/27 & 0/27 & 1/27 \\
$T=2$   & 20/27 & 25/27 & 20/27 \\
\bottomrule
\end{tabular}
\end{table}

Softening lowers MSE in 20 configurations, including one
preference configuration with worse mean calibration.
A reduction in aggregate calibration error therefore
cannot by itself explain the allocation gains.
The temperature intervention changes both calibration and
quota concentration, so their separate contributions
remain unresolved.

\subsection{Information-free score controls}
\label{app:Information-free}
Two negative controls test whether uninformative scores create apparent
stratification gains. We permute scores within original groups 100 times
per configuration, with randomized tie breaking; separately, constant
scores induce random equal-size partitions. Both controls compare
proportional stratified means. Mean stratified/original-group risk ratios
range from $0.9991$ to $1.0014$ for permuted scores and from $0.9991$
to $1.0007$ for constant scores. These results connect the observed
stratification gain to the information in the original score ordering.

\subsection{Additional correction--allocation combinations}
\label{r6:app:allocation_correction}
Table~\ref{r6:tab:extra_combinations} supplies the preference
comparisons omitted from the main results table: fixed correction
($b_h=1$) under both allocations and the unregularized group-shared
fit under working quotas. All use ten bins per natural group and
$L=\lfloor N/4\rfloor$. Fixed-correction risks are exact. Fitted risks
use 10,000 paired draws, 100 independent batches for MCSEs, and the
exact same-quota mean risk as a control variate. Working quotas use
the annotation-specific variance model.

\begin{table}[htbp]
\centering
\caption{Additional correction--allocation comparisons.
Every entry is MSE divided by the matching proportional stratified-mean
risk $R_0$. Fixed correction sets $b_h=1$; Linear fits a group-shared
coefficient. Parentheses give one MCSE. Mean and Ridge results are
reported in the main table. Bold marks ratios outside $[0.95,1.05]$.
M/Q denote Mistral/Qwen.}
\label{r6:tab:extra_combinations}
\small
\begin{tabular}{lrrr}
\toprule
& Proportional & \multicolumn{2}{c}{Working} \\
\cmidrule(lr){2-2}\cmidrule(lr){3-4}
Target / judge & Fixed & Fixed & Linear \\
\midrule
Alpaca / M         & 1.001 & 0.997 & 1.004 (0.0014) \\
Alpaca / Q         & 0.997 & 1.030 & 1.036 (0.0023) \\
Arena mean / M     & 0.999 & 0.997 & 1.005 (0.0018) \\
Arena mean / Q     & 1.001 & 1.039 & 1.042 (0.0016) \\
Arena contrast / M & 0.997 & 1.000 & 1.010 (0.0023) \\
Arena contrast / Q & 1.004 & 1.032 & 1.037 (0.0016) \\
HelpSteer3 / M     & 0.992 & 0.960 & 0.971 (0.0029) \\
HelpSteer3 / Q     & 0.994 & \textbf{0.923} & \textbf{0.930} (0.0020) \\
\bottomrule
\end{tabular}
\end{table}

Fixed correction provides further gains for HelpSteer3 under working
quotas. The remaining preference configurations stay close to, or
above, the proportional-mean reference.

\section{Pilot Reuse: Risk and Interval Calculations}\label{app:r38_pilot}\label{app:pilot-pooling}

\subsection{Risk and variance estimation under pilot reuse}
\label{r6:app:pilot_risk}
We use the sampling designs and estimators defined in
Section~\ref{sec:dif_est}.
Each of the 27 primary configurations uses 2,000 proportional pilots, with
ten bins per natural group and $L=\lfloor N/4\rfloor$.
The matched proportional control uses $q_h^p=n_h^0-m_h$.
Score-only and learned final counts use the bounded allocation routine
with remaining sizes $M_h$ and budget $L-P$. Score-only allocation uses the full-stratum working
variance. Learned allocation uses the pilot residual standard deviation
and falls back to remaining-size proportional allocation if any such
standard deviation is zero. Estimators within a quota rule use the
same observations. For HelpSteer3/Qwen, at least one stratum has zero estimated
pilot-residual standard deviation in every replay, so the learned
rule always uses the remaining-size proportional fallback.
Its Completion, Composite, and Pooling MSE ratios round to
1.19, 1.00, and 1.00, respectively, as do the matched control's ratios.

Let $S^2_{y,h\setminus A}$ denote the variance of the
human outcomes in $U_h\setminus A_h$, with divisor $M_h-1$.
For the single-stratum calculation below, $S_y^2$ denotes
the full-pool human-outcome variance, with divisor $N-1$. Conditional on the pilot $A$, final quotas are fixed and each final
sample mean is unbiased for its remaining-stratum mean. Completion
is therefore conditionally unbiased, with variance
\begin{equation}
 V_C(A)=\sum_h\left(\frac{M_h\omega_h}{N}\right)^2
       \frac{1-q_h/M_h}{q_h}S^2_{y,h\setminus A}.
 \label{r6:eq:completion_variance}
\end{equation}
Its MSE is $\mathbb E_A[V_C(A)]$. We integrate final-stage variation
exactly and average over pilots; paired risk contrasts use the same pilots.
For one SRS stratum, $\mathbb E[S^2_{y,U\setminus A}]=S_y^2$, giving
\begin{equation}
 R_{C,\rm prop}=\frac{(N-P)(N-L)}{N^2(L-P)}S_y^2,
 \qquad R_0=\frac{N-L}{NL}S_y^2,
 \qquad\frac{R_{C,\rm prop}}{R_0}
       =\frac{1-P/N}{1-P/L}.
 \label{r6:eq:completion_cost}
\end{equation}
The ratio also holds across proportional strata with equal sampling
fractions and no integer effects. It describes Completion's weighting.

Because $\mathbb E[C_h\mid A]$ equals the full-stratum mean,
Completion error is uncorrelated with any centered function of the
pilot. Composite consequently has variance
\begin{equation}
\begin{split}
 \operatorname{Var}(\widehat\theta_B)
 &=\sum_h a_h^2\lambda_h^2(1/m_h-1/N_h)S^2_{y,h}\\
 &\quad+\mathbb E_A\!\left[
 \sum_h a_h^2(1-\lambda_h)^2(M_h/N_h)^2
       (1/q_h-1/M_h)S^2_{y,h\setminus A}\right].
\end{split}
\label{r6:eq:composite_variance}
\end{equation}
Pooling's MSE averages its conditional variance plus squared conditional
bias. Its marginal variance is MSE minus squared marginal bias.
For deterministic $m_h,q_h$, the union is an SRS of size $m_h+q_h$:
each possible union has $\binom{m_h+q_h}{m_h}$ equally likely pilot
subsets. Thus matched proportional counts make Composite and Pooling
equal to the one-stage mean, with risk $R_0$. Score-only quotas with
Pooling likewise have the marginal risk of a one-stage mean using
the predetermined counts $m_h+q_h^s$.

Let $s^2_{A,h}$ and $s^2_{T,h}$ be the sample variances
of the human outcomes in $A_h$ and $T_h$, using divisors
$m_h-1$ and $q_h-1$, respectively.
Similarly, $s^2_{A\cup T,h}$ is the human-outcome sample
variance over $A_h\cup T_h$, with divisor $m_h+q_h-1$.
For Completion, replacing the remaining-population variances
by the final sample variances gives a conditionally unbiased
variance estimate.
For Composite, we use
\begin{equation}
\begin{split}
 \widehat V_B
 &=\sum_h a_h^2\lambda_h^2(1/m_h-1/N_h)s^2_{A,h}\\
 &\quad+\sum_h a_h^2(1-\lambda_h)^2(M_h/N_h)^2
       (1/q_h-1/M_h)s^2_{T,h}.
\end{split}
\label{r6:eq:composite_variance_estimate}
\end{equation}
The two terms are unbiased for their counterparts in
Equation~\eqref{r6:eq:composite_variance}.
The ordinary pooled variance estimate is
$\sum_h a_h^2\{1/(m_h+q_h)-1/N_h\}s^2_{A\cup T,h}$.
It is unbiased for fixed quotas; its use under learned quotas is the
interval procedure evaluated in the coverage experiments.
Coverage uses 20 final draws per pilot, with MCSEs clustered by pilot.

\subsection{Explicit shrinkage of learned quotas}\label{app:r40-shrinkage}
\label{r6:app:shrinkage}

Let $q_h^\ell,q_h^p,q_h^s$ denote the learned, proportional, and
score-only final label counts. The mixtures are
\[
 \widetilde q_h^{\ell:p}=\tfrac12(q_h^\ell+q_h^p),
 \qquad
 \widetilde q_h^{\ell:s}=\tfrac12(q_h^\ell+q_h^s).
\]
We floor the averages and assign remaining labels in decreasing order
of fractional remainder, breaking ties stably. The original counts
share the same integer bounds and total, so this rounding preserves
the budget, minima, and capacities. The weight is one half for every
system and was fixed before evaluating either mixture.

All 27 configurations reuse the 2,000 pilots and their 20 final priority
draws. Only final counts change; total budgets and Composite weights
remain fixed. Final-stage MSE is integrated analytically, including
Pooling's squared conditional bias. Composite's pilot-variance term
uses its exact expectation. Paired MCSEs use pilots as independent units.

Every ratio in Table~\ref{tab:r40-shrinkage-all} uses score-only quotas
with Pooling as its reference, including the Completion and Composite
columns. For score-directed shrinkage with Pooling, the AQA exceptions
are Post + reranking (1.010; inconclusive paired contrast) and
RTR + reranking (1.061; favors the reference).
All 17 favorable paired differences exceed two MCSEs.
Proportional-directed shrinkage with Composite beats this reference
only for the two reranking systems, where the proportional mean has
still lower MSE.

\begin{table}[htbp]
\centering\small
\setlength{\tabcolsep}{3pt}
\caption{Shrinkage results for all 27 primary configurations.
Every ratio uses the MSE of score-only quotas with Pooling as its
denominator. Columns compare the three label-reuse estimators
under each shrinkage rule. All procedures use the same pilot
samples and total label budget within a configuration.}
\label{tab:r40-shrinkage-all}
\begin{tabular}{lrrrrrr}
\toprule
& \multicolumn{3}{c}{Toward proportional} & \multicolumn{3}{c}{Toward score-only} \\ \cmidrule(lr){2-4}\cmidrule(lr){5-7} Configuration & Completion & Composite & Pooling & Completion & Composite & Pooling \\
\midrule
Alpaca / M & 1.228 & 1.028 & 1.048 & 1.227 & 1.027 & 1.047 \\
Alpaca / Q & 1.202 & 1.007 & 1.011 & 1.211 & 1.014 & 1.014 \\
Arena contrast / M & 1.275 & 1.062 & 1.089 & 1.275 & 1.063 & 1.088 \\
Arena contrast / Q & 1.192 & 1.000 & 1.005 & 1.193 & 1.003 & 1.001 \\
Arena mean / M & 1.229 & 1.031 & 1.070 & 1.230 & 1.031 & 1.070 \\
Arena mean / Q & 1.174 & 0.988 & 0.991 & 1.169 & 0.986 & 0.986 \\
HelpSteer3 / M & 1.332 & 1.103 & 1.111 & 1.332 & 1.102 & 1.065 \\
HelpSteer3 / Q & 1.294 & 1.085 & 1.086 & 1.210 & 1.027 & 1.024 \\
LLM-as-retriever & 1.238 & 1.066 & 0.989 & 1.422 & 1.196 & 0.891 \\
Post + rerank & 0.987 & 0.826 & 0.836 & 1.385 & 1.109 & 1.010 \\
Post-1 & 1.230 & 1.058 & 1.007 & 1.367 & 1.155 & 0.915 \\
Post-2 & 1.246 & 1.061 & 1.016 & 1.523 & 1.257 & 0.936 \\
Post-3 & 1.231 & 1.054 & 0.993 & 1.461 & 1.217 & 0.912 \\
Post-4 & 1.212 & 1.030 & 0.967 & 1.456 & 1.203 & 0.924 \\
Post-5 & 1.434 & 1.248 & 1.156 & 1.375 & 1.206 & 0.958 \\
Post-6 & 1.233 & 1.048 & 0.987 & 1.552 & 1.274 & 0.910 \\
Post-7 & 1.433 & 1.255 & 1.165 & 1.371 & 1.212 & 0.946 \\
Post-8 & 1.322 & 1.129 & 1.071 & 1.464 & 1.230 & 0.945 \\
RTR + rerank & 1.108 & 0.923 & 0.922 & 1.565 & 1.247 & 1.061 \\
RTR-10 & 1.278 & 1.086 & 1.057 & 1.461 & 1.216 & 0.984 \\
RTR-11 & 1.328 & 1.145 & 1.057 & 1.575 & 1.321 & 0.897 \\
RTR-12 & 1.281 & 1.089 & 1.054 & 1.423 & 1.190 & 0.975 \\
RTR-2 & 1.542 & 1.413 & 1.259 & 1.421 & 1.328 & 0.919 \\
RTR-4 & 1.186 & 1.032 & 0.995 & 1.292 & 1.107 & 0.866 \\
RTR-6 & 1.299 & 1.118 & 1.054 & 1.481 & 1.247 & 0.926 \\
RTR-8 & 1.241 & 1.056 & 1.000 & 1.556 & 1.279 & 0.906 \\
RTR-9 & 1.340 & 1.158 & 1.131 & 1.365 & 1.175 & 0.979 \\
\bottomrule
\end{tabular}
\end{table}

\subsection{Coverage-diagnostic protocol}\label{app:r40-coverage}\label{r6:app:coverage_diagnostic}

Estimating uncertainty and achieving the intended coverage
are separate requirements. Predictive-inference methods
make this distinction explicit by calibrating fitted
predictions or intervals to meet a coverage target
\citep{lei2018distributionfree,romano2019conformalized,
zou2026conformalized}.
For the finite-pool mean intervals studied here, we
diagnose whether undercoverage comes from estimated
widths or biased point estimates.

For each saved point estimate, we first replace the
estimated variance by a fixed reference $V_*$ for that
system and procedure. We then keep this width and subtract
the measured bias from the point estimate. The first comparison changes interval widths while keeping the point
estimates fixed; the second assesses the additional effect of centering
those estimates at the target.

The reference is analytic for fixed allocations. For learned
allocations, we integrate final-stage variation and average over
pilots. Composite adds its analytic pilot-contribution variance;
Pooling uses marginal MSE minus squared marginal bias.
The reference requires complete outcomes and is a retrospective
diagnostic. Paired risk contrasts use common pilots with separately
centered control variates, and coverage MCSEs cluster the 20 final
draws within each of the 2,000 pilots.

For AQA learned Composite, the reference variance $V_*$ has relative
MCSE 0.26--0.66\%.
The coverage MCSEs treat this estimated reference as fixed and cluster
final draws by pilot; they do not include uncertainty from estimating
$V_*$.
As a separate sensitivity check, we vary $V_*$ by
$\pm z_{0.975}$ times its MCSE.
Across systems and the two endpoints, coverage ranges from
93.90\% to 95.28\%, compared with 93.97--95.20\% at the central reference.
The improvement after replacing the sample-based variance therefore
persists over this range of reference variances.

\section{Active-Method Implementation and Checks}\label{app:r38_active}

\subsection{Estimators and acquisition rules}\label{app:active}
We write the target as
$\theta=N^{-1}\sum_{i\in U}\omega_i y_i$, where
$\omega_i=Na_h/N_h$ for $i\in U_h$.
These fixed target weights are distinct from acquisition
probabilities.
Preference comparisons apply each procedure within the
original groups; AQA has one group.
Let $L_g$ be the label budget for group $G_g$, and let
$\omega_g$ be its common item-level target weight, so
$\omega_i=\omega_g$ for $i\in G_g$.
Equivalently, $\omega_g=Nw_g/N_g$, where $w_g$ is the
weight assigned to the group mean.
The group budgets sum to $L$ and are allocated in proportion
to $N_g|\omega_g|$, subject to integer and capacity bounds.

\paragraph{Capped ASI.}
We use GLIDE's implementation~\citep{martinon2026glide} of
ASI~\citep{zrnic2024active}.
Within group $g$, its native proxy is
$v_i=\max\{0.05,|\omega_i|\sqrt{s_i(1-s_i)}\}$.
It chooses $\pi_i=\min\{1,c_gv_i\}$ with
$\sum_{i\in g}\pi_i=L_g$.
After randomly ordering the group and drawing
$\xi_i\sim\operatorname{Bernoulli}(\pi_i)$, it retains the prefix $J_g$
ending before the $(L_g+1)$st positive draw, or the entire group if
that draw does not occur.
Only retained positive draws receive labels; $J_g$ also contains
unlabeled items. The estimate is
\begin{equation}
 \widehat\theta_A=\sum_g\frac{N_g}{N}\overline z_g(\widehat b_g),
 \qquad
 z_i(b)=\omega_i\left\{bs_i+\frac{\xi_i}{\pi_i}(y_i-bs_i)\right\},
 \qquad
 \overline z_g(b)=\frac1{|J_g|}\sum_{i\in J_g}z_i(b).
 \label{eq:capped_asi_estimator_main}
\end{equation}
The fitted $\widehat b_g$ minimizes the empirical variance of $z_i(b)$
over $J_g$, using the acquired labels without an additional pilot.
The probabilities in the correction are the Bernoulli probabilities
before the cap. 

\paragraph{RobustASI.}
We use the published implementation~\citep{li2025robust}, with a
paid pilot $A$ of size $P$ and independent Bernoulli acquisition from
$U\setminus A$. If $T$ is the acquired final sample, our finite-pool
estimate is
\begin{equation}
 \widehat\theta_R=\frac1N\left\{
 \sum_{i\in U}\omega_i s_i+
 \sum_{i\in A}\omega_i(y_i-s_i)+
 \sum_{i\in T}\frac{\omega_i(y_i-s_i)}{\pi_i}\right\}.
 \label{eq:robust_estimator_main}
\end{equation}
This adaptation retains the fixed prediction coefficient $b=1$ in
the RobustASI mean estimator, whereas capped ASI fits its coefficient.
The same-pilot uniform control retains $b=1$ as well, so their
comparison isolates the acquisition rule within this estimator.
For one group, let $B=L-P$, $\bar\pi=B/(N-P)$,
$u_i^{\rm act}=1-2|s_i-1/2|$, and $\bar u^{\rm act}$ be its mean outside the pilot.
The native rule interpolates between uncertainty-based and uniform
sampling:
\[
 p_i=\min\{1,\bar\pi u_i^{\rm act}/\bar u^{\rm act}\},\qquad
 g_i=p_i^{1-\widehat\rho}\bar\pi^{\widehat\rho},\qquad
 \pi_i=\min\left\{1,\frac{Bg_i}{\sum_{j\notin A}g_j}\right\}.
\]
The endpoint $\widehat\rho=1$ is uniform sampling.
The error model predicts $|y-s|$ from $\max(s,1-s)$ using XGBoost
(100 trees, learning rate 0.1, depth six).
These predictions enter the robust objective that selects
$\widehat\rho$ on a 100-point grid.
Five-fold pilot cross-validation selects the radius from
$\{0.02,0.03,0.04,0.05,0.06,0.07\}$, with 200 validation draws per fold.
Pilot totals are 32 labels for AQA and Alpaca and 64 for Arena and
HelpSteer3. All fitting uses those paid labels.

Positive remaining probabilities make Equation~\eqref{eq:robust_estimator_main}
conditionally unbiased. For zero-probability items, the uncorrected
residual contributes bias. With $e_i=\omega_i(y_i-s_i)$ and
$Z=\{i\notin A:\pi_i=0\}$, the conditional variance and bias are
\begin{equation}
 V(A)=\frac1{N^2}\sum_{\substack{i\notin A\\\pi_i>0}}
 e_i^2(\pi_i^{-1}-1),\qquad
 B(A)=-\frac1N\sum_{i\in Z}e_i.
 \label{eq:bernoulli}
\end{equation}
MSE averages $V(A)+B(A)^2$ over pilots.
We sum acquired residuals only and keep the zero-support bias;
probabilities are not modified after acquisition.
The uniform control uses the same pilot and estimator with
$\pi_i=\bar\pi$.

\subsection{Results and actual label expenditure}\label{app:active_results}\label{r6:app:active_results}
Tables~\ref{tab:active_errors} and~\ref{tab:pref_active} report the
AQA and preference comparisons for capped ASI and RobustASI
\citep{zrnic2024active,li2025robust}.
Let $K$ include every acquired label, including the pilot, and let
$R_{\rm prop}(k)$ be the exact risk of a fresh proportional stratified
mean using $k$ labels and the same bounded allocation routine.
We report
\[
 \frac{\operatorname{MSE}(\widehat\theta)}{R_{\rm prop}(L)}
 \qquad\text{and}\qquad
 \frac{\operatorname{MSE}(\widehat\theta)}
      {\mathbb E_K[R_{\rm prop}(K)]}.
\]
The second denominator averages reference risks over the method's
realized count distribution. It adjusts the comparison for expenditure;
it does not impose a cap on the active sampler.

At nominal $L=250$ on AQA, capped ASI averages 244.904--245.589 labels
across systems. RobustASI averages 250.087--250.189 in simulation and
satisfies its conditional expected budget for every pilot, but exceeds
250 labels in 48.135--48.730\% of runs. The numbers of systems favoring
capped ASI and RobustASI remain seven and four under either reference.
On HelpSteer3/Qwen, the capped ASI ratio changes from
$0.991\pm0.0047$ to $0.951\pm0.0045$ (one MCSE), so expenditure
adjustment matters for this comparison.

\begin{table}[htbp]
\centering\small
\setlength{\tabcolsep}{3.5pt}
\caption{AQA active comparisons. Nominal ratios use $R_{\rm prop}(L)$; actual-$K$ ratios use $\mathbb E_K[R_{\rm prop}(K)]$. The last column uses the same-pilot uniform control. Reported errors are one MCSE.}
\label{tab:active_errors}
\begin{tabular}{lrrrrr}
\toprule
System & ASI nominal & Actual $K$ & Robust nominal & Actual $K$ & Robust/uniform \\
\midrule
LLM-as-retriever & 1.066 $\pm$ 0.005 & 1.036 & 1.025 $\pm$ 0.004 & 1.019 & 0.850 $\pm$ 0.003 \\
Post+rerank & 1.651 $\pm$ 0.007 & 1.604 & 1.353 $\pm$ 0.006 & 1.347 & 1.028 $\pm$ 0.004 \\
Post-1 & 1.131 $\pm$ 0.005 & 1.099 & 1.049 $\pm$ 0.004 & 1.043 & 0.866 $\pm$ 0.003 \\
Post-2 & 1.157 $\pm$ 0.005 & 1.125 & 1.056 $\pm$ 0.003 & 1.051 & 0.882 $\pm$ 0.003 \\
Post-3 & 1.173 $\pm$ 0.006 & 1.140 & 1.076 $\pm$ 0.004 & 1.070 & 0.872 $\pm$ 0.003 \\
Post-4 & 1.253 $\pm$ 0.006 & 1.218 & 1.109 $\pm$ 0.003 & 1.104 & 0.898 $\pm$ 0.003 \\
Post-5 & 0.760 $\pm$ 0.004 & 0.738 & 0.913 $\pm$ 0.004 & 0.908 & 0.805 $\pm$ 0.003 \\
Post-6 & 1.160 $\pm$ 0.005 & 1.128 & 1.083 $\pm$ 0.003 & 1.081 & 0.893 $\pm$ 0.003 \\
Post-7 & 0.737 $\pm$ 0.003 & 0.716 & 0.909 $\pm$ 0.005 & 0.903 & 0.756 $\pm$ 0.004 \\
Post-8 & 0.919 $\pm$ 0.004 & 0.894 & 1.010 $\pm$ 0.002 & 1.007 & 0.901 $\pm$ 0.002 \\
RTR+rerank & 1.443 $\pm$ 0.006 & 1.406 & 1.183 $\pm$ 0.001 & 1.184 & 0.980 $\pm$ 0.001 \\
RTR-10 & 1.035 $\pm$ 0.004 & 1.009 & 1.091 $\pm$ 0.003 & 1.093 & 0.902 $\pm$ 0.002 \\
RTR-11 & 0.949 $\pm$ 0.004 & 0.923 & 0.987 $\pm$ 0.006 & 0.981 & 0.774 $\pm$ 0.004 \\
RTR-12 & 1.040 $\pm$ 0.005 & 1.013 & 1.050 $\pm$ 0.003 & 1.050 & 0.874 $\pm$ 0.003 \\
RTR-2 & 0.731 $\pm$ 0.004 & 0.708 & 0.829 $\pm$ 0.006 & 0.816 & 0.715 $\pm$ 0.005 \\
RTR-4 & 1.253 $\pm$ 0.006 & 1.217 & 1.098 $\pm$ 0.004 & 1.092 & 0.840 $\pm$ 0.003 \\
RTR-6 & 0.961 $\pm$ 0.004 & 0.935 & 1.024 $\pm$ 0.003 & 1.021 & 0.847 $\pm$ 0.003 \\
RTR-8 & 1.132 $\pm$ 0.006 & 1.103 & 1.115 $\pm$ 0.003 & 1.115 & 0.899 $\pm$ 0.002 \\
RTR-9 & 0.926 $\pm$ 0.004 & 0.902 & 1.046 $\pm$ 0.003 & 1.045 & 0.840 $\pm$ 0.002 \\
\bottomrule
\end{tabular}
\end{table}

\begin{table}[htbp]
\centering\small
\setlength{\tabcolsep}{3.5pt}
\caption{Preference comparisons with original groups respected. Denominators match Table~\ref{tab:active_errors}; M/Q denote Mistral/Qwen. Errors are one MCSE. RobustASI MSE includes zero-support residual bias.}
\label{tab:pref_active}
\begin{tabular}{lrrrrr}
\toprule
Target / score & ASI nominal & Actual $K$ & Robust nominal & Actual $K$ & Robust/uniform \\
\midrule
Alpaca / M & 1.101 & 1.060 $\pm$ 0.005 & 1.246 & 1.242 $\pm$ 0.001 & 1.000 $\pm$ 0.000 \\
Alpaca / Q & 1.254 & 1.208 $\pm$ 0.005 & 1.308 & 1.304 $\pm$ 0.002 & 1.005 $\pm$ 0.001 \\
Arena mean / M & 1.046 & 1.017 $\pm$ 0.005 & 1.181 & 1.181 $\pm$ 0.000 & 1.000 $\pm$ 0.000 \\
Arena mean / Q & 1.165 & 1.133 $\pm$ 0.005 & 1.175 & 1.175 $\pm$ 0.001 & 1.004 $\pm$ 0.001 \\
Arena contrast / M & 1.049 & 1.009 $\pm$ 0.005 & 1.312 & 1.309 $\pm$ 0.001 & 1.000 $\pm$ 0.000 \\
Arena contrast / Q & 1.211 & 1.167 $\pm$ 0.005 & 1.320 & 1.318 $\pm$ 0.002 & 1.008 $\pm$ 0.001 \\
HelpSteer3 / M & 1.051 & 1.011 $\pm$ 0.004 & 1.084 & 1.083 $\pm$ 0.001 & 0.963 $\pm$ 0.001 \\
HelpSteer3 / Q & 0.991 & 0.951 $\pm$ 0.005 & 1.059 & 1.056 $\pm$ 0.001 & 0.945 $\pm$ 0.001 \\
\bottomrule
\end{tabular}
\end{table}

\FloatBarrier

\paragraph{Error-model sensitivity.}
A score that predicts the mean outcome need not describe
the variability around it. Distributional regression
models this additional information by allowing scale and
other features of the outcome distribution to depend on
the inputs \citep{rigby2005gamlss,duan2020ngboost,
hu2026conditional}.
We test whether changing the error-model inputs changes
the active procedure's advantage over stratification. We replace RobustASI's folded-confidence input $x=\max(s,1-s)$ by
$s$ or $(s,x)$, retaining all other settings. A fourth fit regresses
$(y-s)^2$ on $(1,z,z^2)$, $z=2s-1$, with ridge penalty one on the
nonintercept coefficients, clips predictions to $[0,1]$, and takes
their square root. Using 100 common pilots and 100 paired final draws
per pilot, the folded, raw, and combined inputs each beat the nominal
proportional stratified reference for four of 19 AQA systems; the
quadratic fit does so for two. None of the eight preference
configurations beats that reference. The tested changes leave the
comparison with stratification qualitatively unchanged.

\subsection{Selective acquisition within score strata}
We also test whether active acquisition adds value after stratification.
Within each stratum, LURE~\citep{farquhar2021bias} samples without
replacement and applies its unbiased sequential weights, using exactly
the proportional or working quota.
One estimator applies LURE to $y$; the other adds the full-score mean
to LURE applied to $y-s$.
Within stratum $h$, proposal weights are proportional to
$0.1+0.9d_i/\max_{j\in U_h}d_j$, with uniform sampling
when all $d_i=0$.
Using the working mean $s_i$ and variance $u_i$ from the
allocation comparison, we set $d_i=\sqrt{u_i+s_i^2}$ for
outcomes and $d_i=\sqrt{u_i}$ for residuals.
These are fixed-surrogate controls; prediction-powered active testing
additionally fits a prediction coefficient~\citep{ashouritaklimi2026ppat}.
All 27 configurations use 10,000 paired trials and the same-quota
uniform mean's loss as a control variate with known expectation.
Table~\ref{tab:r45_lure} reports MSE ratios to that same-quota mean.
The ranges straddle one: selective acquisition gives some improvements,
but does not consistently improve on uniform sampling within strata.

\begin{table}[htbp]
\centering\small
\setlength{\tabcolsep}{4pt}
\caption{Fixed-budget active-testing controls. MSE and its MCSE both use the same-quota uniform mean as reference. Working stratum quotas and within-stratum second-moment proposals follow the annotation-specific model.}
\label{tab:r45_lure}
\begin{tabular}{lllrr}
\toprule
Scope & Quota & Estimator & MSE ratio & MCSE \\
\midrule
AQA & Proportional & LURE & 0.952--1.007 & 0.0047--0.0088 \\
AQA & Proportional & Residual LURE & 0.953--1.159 & 0.0075--0.0142 \\
AQA & Working & LURE & 0.926--1.757 & 0.0026--0.1532 \\
AQA & Working & Residual LURE & 0.908--1.746 & 0.0079--0.1538 \\
Preference & Proportional & LURE & 0.988--1.022 & 0.0030--0.0056 \\
Preference & Proportional & Residual LURE & 0.996--1.036 & 0.0022--0.0077 \\
Preference & Working & LURE & 0.987--1.043 & 0.0030--0.0063 \\
Preference & Working & Residual LURE & 0.995--1.056 & 0.0022--0.0083 \\
\bottomrule
\end{tabular}
\end{table}

\section{Ranking and Small-Budget Sensitivities}
\subsection{Coupling, metrics, and Monte Carlo precision}\label{app:ranking}
Limited observations are often used to support a final
ranking or voting decision, so the value of an additional
observation depends on its contribution to that decision
\citep{psr8_chatzi_ranking,aggarwal2023let,
gui2026answers}.
We therefore evaluate the decision itself across budgets.
For our leaderboard, we also vary how samples are
coordinated across systems.
The pool contains the same 19 AQA systems on 1,000 questions;
each system uses ten score strata and $L$ human labels.
The main design shares question priorities across systems,
whereas the sensitivity uses independent priorities for
each system. Within either joint design, proportional and
working quotas are compared with paired randomness at a
total cost of $19L$ labels.

At $L=150,250,350$, the shared-priority analysis uses
100,000 joint draws.
To obtain independent-system samples, we assign each of the 19 systems
a different block of 5,000 draws from these saved results.
Combining the first estimate from each block gives one complete
leaderboard, and repeating this construction gives 5,000 independent
leaderboards.
At $L=50,100,200$, both joint designs use 20,000 paired replays.
Metrics and counts were fixed before calculating the
ranking extension. These static draws differ from the pilot panel in
Table~\ref{tab:ranking_r38}.

The 19 full-pool means are distinct, giving 171 pairwise orderings.
An estimated tie counts as half an inversion, with tolerance
$10^{-12}$. Best-system selection and top-three recall average
uniformly over estimated ties. For the static comparisons, MCSEs use complete ranking
replays as independent replications.
The pilot panel in Table~\ref{tab:ranking_r38} uses
2,000 independent pilot replicates, each with 20
final-stage draws.
For that panel, we first average each ranking metric over
the 20 final draws within a pilot, then compute its standard
error across the 2,000 pilot averages.
Neither system pairs nor systems are treated as independent
replications. Contrasts pair working and proportional sampling
within each replay.

The table reports pairwise inversion under both joint sampling designs.
At $L=250$ with shared priorities, average pairwise error variance
rises from $0.00053121$ to $0.00059283$ under working quotas; the
covariance change accounts for about 80\% of this increase.
The full-pool winner is RTR + rerank, with human mean $0.714$,
followed by RTR-10 at $0.655$ and RTR-12 at $0.633$.
For these two comparisons with the winner, covariance changes account
for approximately 9\% and 22\% of the increase in difference-estimation
variance, respectively. The remaining increase comes from the
individual variances. Under independent sampling at the same budget,
proportional and working quotas select the best system in 98.04\%
and 96.04\% of replays, respectively. Thus the slight improvement in
average pairwise ordering under independent sampling does not extend
to best-system selection.
\begin{table}[t]
\centering\small
\caption{AQA pairwise inversion rates under both joint sampling designs. Rates are percentages; $\Delta$ is working minus proportional in percentage points, with paired MCSE in parentheses. Lower inversion is better. A ranking costs $19L$ system--question labels.}
\label{tab:ranking-designs}
\begin{tabular}{r rrr rrr}
\toprule
& \multicolumn{3}{c}{Shared question priorities} & \multicolumn{3}{c}{Independent systems} \\
\cmidrule(lr){2-4}\cmidrule(lr){5-7}
$L$ & Prop. & Working & $\Delta$ (MCSE) & Prop. & Working & $\Delta$ (MCSE) \\
\midrule
50 & 15.256 & 15.109 & -0.146 (0.029) & 16.756 & 16.388 & -0.369 (0.026) \\
100 & 11.108 & 11.270 & +0.162 (0.024) & 12.531 & 12.413 & -0.118 (0.023) \\
150 & 9.051 & 9.211 & +0.160 (0.010) & 10.356 & 10.179 & -0.177 (0.042) \\
200 & 7.672 & 7.810 & +0.138 (0.019) & 8.905 & 8.718 & -0.187 (0.019) \\
250 & 6.683 & 6.832 & +0.149 (0.008) & 7.820 & 7.660 & -0.161 (0.036) \\
350 & 5.237 & 5.376 & +0.139 (0.007) & 6.249 & 6.071 & -0.178 (0.032) \\
\bottomrule
\end{tabular}
\end{table}

\subsection{Small budgets and a five-percent sampling fraction}\label{app:small_budget}
The 27 target--score configurations use $L=50,100,200$ and
$L=\lfloor0.05N\rfloor$, computing coincident totals once. Within each
original group, 2, 5, or 10 score bins are crossed with proportional
and working quotas. Mean MSE is exact. Local and group-shared fits in the proportional-quota sensitivity
calculations use 20,000 paired draws and 200 independent batches for
MCSEs. In the working-quota fitted column of
Table~\ref{tab:small-budget-primary}, preference configurations use
10,000 paired draws and 100 batches, whereas AQA uses 20,000 draws
and 200 batches. The squared estimation error of the same-quota
stratified mean, whose expectation is known exactly, serves as a
control variate for estimating fitted MSEs.

Table~\ref{tab:small-budget-primary} uses the original-group
proportional mean at the same budget as its denominator, so bin counts
can be compared directly. For HelpSteer3/Qwen at $L=100$, five-bin and
ten-bin proportional means have ratios 0.623 and 0.640. The latter
places 30 of 40 strata at two labels. At five percent, ten-bin local
PPI has 1.304--2.056 times its matching mean risk on preference pools
and 1.140--1.202 on AQA; these within-partition ratios have a different
denominator from the table. Sharing coefficients is more stable.
The working-quota improvement for HelpSteer3/Qwen falls from 7.13\%
at the primary budget to 2.57\% at five percent. Floors, integer
rounding, and finite-population terms all change in this comparison.
\begin{table}[htbp]
\centering\small
\caption{Five-percent budgets: every entry is MSE divided by the exact original-group proportional-mean MSE at the same $L=\lfloor0.05N\rfloor$. This single denominator permits comparisons across partitions. The first three numeric risk columns use proportional means; the last two use ten-bin working quotas with a mean or group-shared PPI fit. Mean risks are exact. Parentheses give PPI MCSEs; preference fits use 10,000 draws and AQA fits use 20,000. Working quotas use the annotation-specific model. M/Q denote Mistral/Qwen, both using order-averaged scores. Rerank denotes AutoAIS reranking.}
\label{tab:small-budget-primary}
\begin{tabular}{lr rrr rr}
\toprule
& & \multicolumn{3}{c}{Proportional mean} & \multicolumn{2}{c}{Working, ten bins} \\
\cmidrule(lr){3-5}\cmidrule(lr){6-7}
Target / score & $L$ & 2 bins & 5 bins & 10 bins & Mean & Shared PPI \\
\midrule
Alpaca / M & 32 & 0.871 & 0.823 & 0.828 & 0.817 & 0.882 (0.0053) \\
Alpaca / Q & 32 & 0.841 & 0.736 & 0.725 & 0.735 & 0.782 (0.0040) \\
Arena mean / M & 143 & 0.948 & 0.934 & 0.936 & 0.940 & 1.007 (0.0053) \\
Arena mean / Q & 143 & 0.907 & 0.871 & 0.863 & 0.881 & 0.905 (0.0030) \\
Arena contrast / M & 62 & 0.947 & 0.940 & 0.938 & 0.939 & 1.025 (0.0068) \\
Arena contrast / Q & 62 & 0.923 & 0.878 & 0.863 & 0.897 & 0.954 (0.0045) \\
HelpSteer3 / M & 100 & 0.757 & 0.714 & 0.730 & 0.730 & 0.779 (0.0037) \\
HelpSteer3 / Q & 100 & 0.697 & 0.623 & 0.640 & 0.624 & 0.660 (0.0029) \\
\midrule
LLM-as-retriever & 50 & 0.490 & 0.430 & 0.415 & 0.374 & 0.375 (0.0008) \\
Post rerank & 50 & 0.717 & 0.608 & 0.598 & 0.768 & 0.768 (0.0009) \\
Post-1 & 50 & 0.504 & 0.471 & 0.443 & 0.428 & 0.432 (0.0008) \\
Post-2 & 50 & 0.546 & 0.501 & 0.481 & 0.476 & 0.479 (0.0009) \\
Post-3 & 50 & 0.506 & 0.462 & 0.442 & 0.430 & 0.435 (0.0009) \\
Post-4 & 50 & 0.516 & 0.482 & 0.457 & 0.462 & 0.467 (0.0008) \\
Post-5 & 50 & 0.423 & 0.375 & 0.356 & 0.282 & 0.281 (0.0007) \\
Post-6 & 50 & 0.575 & 0.493 & 0.483 & 0.484 & 0.490 (0.0008) \\
Post-7 & 50 & 0.384 & 0.373 & 0.338 & 0.255 & 0.260 (0.0005) \\
Post-8 & 50 & 0.529 & 0.468 & 0.456 & 0.411 & 0.410 (0.0010) \\
RTR rerank & 50 & 0.839 & 0.726 & 0.676 & 0.763 & 0.768 (0.0014) \\
RTR-10 & 50 & 0.704 & 0.610 & 0.574 & 0.542 & 0.548 (0.0012) \\
RTR-11 & 50 & 0.516 & 0.391 & 0.388 & 0.349 & 0.350 (0.0008) \\
RTR-12 & 50 & 0.670 & 0.569 & 0.546 & 0.515 & 0.515 (0.0011) \\
RTR-2 & 50 & 0.649 & 0.403 & 0.365 & 0.259 & 0.260 (0.0005) \\
RTR-4 & 50 & 0.557 & 0.471 & 0.454 & 0.481 & 0.481 (0.0010) \\
RTR-6 & 50 & 0.573 & 0.454 & 0.445 & 0.418 & 0.418 (0.0010) \\
RTR-8 & 50 & 0.636 & 0.538 & 0.526 & 0.519 & 0.519 (0.0010) \\
RTR-9 & 50 & 0.627 & 0.492 & 0.475 & 0.407 & 0.410 (0.0008) \\
\bottomrule
\end{tabular}
\end{table}

The tested two-label floor requires $L\geq2H$ for a static design
and $L\geq4H$ for two waves, with adequate remaining capacity in
each stratum. At $L=50$, AQA's minimum pilot uses 20 labels,
or 40\% of the budget. HelpSteer3 with ten bins per natural group
is infeasible statically at $L=50$ and for the two-wave procedure
at $L=100$.

At five-percent budgets, finite-population correction reduces ten-bin
proportional-mean variance by 4.77--5.13\%. For a fixed partition it
subtracts the same $\sum_h a_h^2S_{y,h}^2/N_h$ under every allocation,
so it cannot reverse the working-versus-proportional ordering. These
results use smaller sampling fractions in the observed pools.

\section{Budget and Scoring-Cost Accounting}\label{app:cost}

\subsection{Scoring resources}
Table~\ref{tab:score_cost} includes both response orders and
logged failed calls. Missing failed-call usage remains unknown.
The protocols differ in model, quantization, hardware, and output
format, so the reported times characterize these runs rather than a
common unit of computing cost.
\begin{table}[htbp]
\centering\small
\caption{Recorded resources for the both-order preference scores.
Input counts exclude unobserved failed-call tokens and, for Mistral,
batch padding. Zero Mistral output tokens reflects forward-only likelihood
scoring. Times have the measurement units shown and should not be summed
as a common hardware-cost quantity.}
\label{tab:score_cost}
\begin{tabular}{llrrrrl}
\toprule
Pool & Protocol & Valid & Failed & Input tokens & Output tokens & Recorded time \\
\midrule
Alpaca & Qwen & 1,296 & 2 & 545,495 & 48,784 & 5.432 h, request sum \\
Arena & Qwen & 5,740 & 0 & 5,921,032 & 217,142 & 48.546 h, request sum \\
HelpSteer3 & Qwen & 4,034 & 11 & 6,449,542 & 148,258 & 5.335 h, request sum \\
Alpaca & Mistral & 1,296 & 0 & 497,730 & 0 & 68.865 s, batch share \\
Arena & Mistral & 5,740 & 0 & 5,655,830 & 0 & 800.265 s, batch share \\
HelpSteer3 & Mistral & 4,034 & 0 & 6,586,190 & 0 & 947.061 s, batch share \\
\bottomrule
\end{tabular}
\end{table}

The Qwen runs for Alpaca and Arena used an Apple M2 host with 24\,GiB
of unified memory. Their recorded times sum the elapsed durations of
individual inference requests, including recorded failed attempts.
Requests may overlap, so these sums are not elapsed wall-clock time
or allocated accelerator time. Mistral's Alpaca and Arena runs share one allocation of 1,854
GPU-seconds; dataset-specific allocation times were not measured.
Mistral HelpSteer3 uses 1,971 GPU-seconds.
Qwen HelpSteer3 uses 6,058 GPU-seconds across five jobs, including
startup, failed requests, and continuation. Concurrent requests make
summed request time different from allocated time. Eleven failed
requests lack token metadata, so the reported token usage is a lower
bound. The M2 runs lack a reliable monetary ledger.
AQA requires no new scoring here; the original production cost is
unknown. Arena's mean and contrast share their scores, as do
estimators using the same pool.

\subsection{From resource counts to total cost}
\label{r6:app:total_cost}
For one deployment, let $c_H$ be the price of one released aggregate
outcome and let $M$ be the number of real deployments sharing the
same score pool.
The cost per deployment is
\begin{equation}
 C=c_HK+C_{\rm score}/M+C_{\rm fit}+C_{\rm prepare}.
 \label{r6:eq:total_cost}
\end{equation}
Here, $K$ includes pilot and final labels.
$C_{\rm score}$ is the cost of producing the shared score
pool, while $C_{\rm fit}$ and $C_{\rm prepare}$ are the
fitting and preparation costs for one deployment.
All terms are expressed in the same monetary unit. Equal outcome counts need
not imply equal human labor across datasets because aggregation
protocols differ. Monte Carlo repetitions do not amortize score
production. Existing-score users incur no new score-production cost;
otherwise that cost reduces the human labels affordable under a
fixed financial budget.
Inference expenditure can also be reduced by adapting
which models are queried \citep{chen2024frugalgpt} or
when a reward model performs explicit reasoning
\citep{xue-etal-2026-reason}.
Our comparisons hold the available scores fixed and
assess their use with human labels. Equal label
budgets imply equal total cost only when the other costs are shared
or explicitly included.

\end{document}